\documentclass[utf8]{article}
\usepackage[utf8]{inputenc}

\usepackage{amsmath,amssymb,amsthm,bm,mathtools,multirow}
\usepackage{placeins}
\usepackage{array}
\usepackage{authblk}
\usepackage{algorithm,algorithmic}
\usepackage[english]{babel}
\usepackage{url,hyperref,lineno,microtype,subcaption}

\usepackage[disable]{todonotes} %

\DeclareMathOperator*{\argmin}{arg\,min}

\newcommand{\T}{\operatorname{T}}

\providecommand{\theoremname}{Theorem}

\providecommand{\corollaryname}{Corollary}

\providecommand{\definitionname}{Definition}

\newcommand{\Cat}[1]{\mathcal{C}at\!\left({#1}\right)}

\newcommand{\E}[2]{\mathbb{E}_{#1}\!\left[{#2}\right]}
\newcommand{\U}[2]{\operatorname{U}_{#1}\!\left[{#2}\right]}
\newcommand{\B}{\operatorname{B}}
\newcommand{\F}{\operatorname{F}}
\newcommand{\G}{\operatorname{G}}
\renewcommand{\H}[1]{\operatorname{H}\!\left[{#1}\right]}
\newcommand{\HB}[1]{\operatorname{H}_{\operatorname{B}}\!\left[{#1}\right]}
\newcommand{\KL}[2]{\operatorname{KL}\!\left[{#1}\|{#2}\right]}

\definecolor{myred}{rgb}{0.7, 0.0328, 0.0539}

\usepackage{tikz,colortbl}
\usepackage{xifthen}
\usetikzlibrary{intersections}
\usetikzlibrary{calc}
\usepackage{zref-savepos}
\usepackage{tkz-euclide}

\pgfdeclarelayer{bg}    
\pgfsetlayers{bg,main}  
\definecolor{beige}{RGB}{245, 245, 220}
\definecolor{darkgrey}{RGB}{65, 65, 65}
\definecolor{grey}{RGB}{240, 240, 240}
\definecolor{lightgrey}{RGB}{250, 250, 250}

\newcommand*\smallcircled[1]{\tikz[baseline=(char.base)]{
            \node[shape=circle,draw,inner sep=0pt, text width=3mm, align=center, minimum width=3.5mm] (char) {#1};}}
\newcommand*\smalldarkcircled[1]{\tikz[baseline=(char.base)]{
            \node[shape=circle,draw,inner sep=0pt, text width=3mm, align=center, minimum width=3.5mm, fill=darkgrey, text=white, font=\bfseries] (char) {#1};}}

\newcommand*{\notableentry}{\cellcolor{grey}}

\usetikzlibrary{calc, arrows, fit, positioning, patterns, decorations.pathreplacing, shapes}
\tikzstyle{dash} = [dashed]
\tikzstyle{line} = [draw]
\tikzstyle{box} = [draw, minimum size=.8cm]
\tikzstyle{smallbox} = [draw, minimum size=5mm]
\tikzstyle{roundbox} = [draw, circle, inner sep=0pt, minimum size=.8cm]
\tikzstyle{clamped} = [draw, fill=black, minimum size=0.35cm]
\tikzstyle{optim} = [draw, circle, inner sep=0pt, fill=white, minimum size=0.35cm]
\tikzstyle{msgcircle} = [shape=circle, draw, inner sep=0pt, minimum size=5mm, fill=white]
\tikzstyle{darkmsgcircle} = [shape=circle, draw, inner sep=0pt, minimum size=5mm, fill=darkgrey, text=white, font=\bfseries]
\tikzstyle{msgdoublecircle} = [shape=circle, double, double distance=1.5pt, draw, inner sep=0pt, minimum size=5mm, fill=white]
\tikzstyle{darkmsgdoublecircle} = [shape=circle, double, double distance=1.5pt, draw, inner sep=0pt, minimum size=5mm, fill=darkgrey, text=white, font=\bfseries]

\newcommand{\msg}[6]{
      \ifthenelse{\isin{#1}{left} \AND \isin{#2}{down}}{
            \coordinate (anchor) at ($({#3})!{#5}!({#4})$);
            \node[msgcircle, xshift=-5.5mm] at (anchor) {#6};
            \node[xshift=-1.5mm] at (anchor) {$\downarrow$};
      }{}
      \ifthenelse{\isin{#1}{right} \AND \isin{#2}{down}}{
            \coordinate (anchor) at ($({#3})!{#5}!({#4})$);
            \node[msgcircle, xshift=5.5mm] at (anchor) {#6};
            \node[xshift=1.5mm] at (anchor) {$\downarrow$};
      }{}

      \ifthenelse{\isin{#1}{down} \AND \isin{#2}{right}}{
            \coordinate (anchor) at ($({#3})!{#5}!({#4})$);
            \node[msgcircle, yshift=-6.0mm] at (anchor) {#6};
            \node[yshift=-2.0mm] at (anchor) {$\rightarrow$};
      }{}
      \ifthenelse{\isin{#1}{up} \AND \isin{#2}{right}}{
            \coordinate (anchor) at ($({#3})!{#5}!({#4})$);
            \node[msgcircle, yshift=6.0mm] at (anchor) {#6};
            \node[yshift=2.0mm] at (anchor) {$\rightarrow$};
      }{}

      \ifthenelse{\isin{#1}{down} \AND \isin{#2}{left}}{
            \coordinate (anchor) at ($({#3})!{#5}!({#4})$);
            \node[msgcircle, yshift=-6.0mm] at (anchor) {#6};
            \node[yshift=-2.0mm] at (anchor) {$\leftarrow$};
      }{}
      \ifthenelse{\isin{#1}{up} \AND \isin{#2}{left}}{
            \coordinate (anchor) at ($({#3})!{#5}!({#4})$);
            \node[msgcircle, yshift=6.0mm] at (anchor) {#6};
            \node[yshift=2.0mm] at (anchor) {$\leftarrow$};
      }{}

      \ifthenelse{\isin{#1}{left} \AND \isin{#2}{up}}{
            \coordinate (anchor) at ($({#3})!{#5}!({#4})$);
            \node[msgcircle, xshift=-5.5mm] at (anchor) {#6};
            \node[xshift=-1.5mm] at (anchor) {$\uparrow$};
      }{}
      \ifthenelse{\isin{#1}{right} \AND \isin{#2}{up}}{
            \coordinate (anchor) at ($({#3})!{#5}!({#4})$);
            \node[msgcircle, xshift=5.5mm] at (anchor) {#6};
            \node[xshift=1.5mm] at (anchor) {$\uparrow$};
      }{}
}

\newcommand{\darkmsg}[6]{
      \ifthenelse{\isin{#1}{left} \AND \isin{#2}{down}}{
            \coordinate (anchor) at ($({#3})!{#5}!({#4})$);
            \node[darkmsgcircle, xshift=-5.5mm] at (anchor) {#6};
            \node[xshift=-1.5mm] at (anchor) {$\downarrow$};
      }{}
      \ifthenelse{\isin{#1}{right} \AND \isin{#2}{down}}{
            \coordinate (anchor) at ($({#3})!{#5}!({#4})$);
            \node[darkmsgcircle, xshift=5.5mm] at (anchor) {#6};
            \node[xshift=1.5mm] at (anchor) {$\downarrow$};
      }{}

      \ifthenelse{\isin{#1}{down} \AND \isin{#2}{right}}{
            \coordinate (anchor) at ($({#3})!{#5}!({#4})$);
            \node[darkmsgcircle, yshift=-6.0mm] at (anchor) {#6};
            \node[yshift=-2.0mm] at (anchor) {$\rightarrow$};
      }{}
      \ifthenelse{\isin{#1}{up} \AND \isin{#2}{right}}{
            \coordinate (anchor) at ($({#3})!{#5}!({#4})$);
            \node[darkmsgcircle, yshift=6.0mm] at (anchor) {#6};
            \node[yshift=2.0mm] at (anchor) {$\rightarrow$};
      }{}

      \ifthenelse{\isin{#1}{down} \AND \isin{#2}{left}}{
            \coordinate (anchor) at ($({#3})!{#5}!({#4})$);
            \node[darkmsgcircle, yshift=-6.0mm] at (anchor) {#6};
            \node[yshift=-2.0mm] at (anchor) {$\leftarrow$};
      }{}
      \ifthenelse{\isin{#1}{up} \AND \isin{#2}{left}}{
            \coordinate (anchor) at ($({#3})!{#5}!({#4})$);
            \node[darkmsgcircle, yshift=6.0mm] at (anchor) {#6};
            \node[yshift=2.0mm] at (anchor) {$\leftarrow$};
      }{}

      \ifthenelse{\isin{#1}{left} \AND \isin{#2}{up}}{
            \coordinate (anchor) at ($({#3})!{#5}!({#4})$);
            \node[darkmsgcircle, xshift=-5.5mm] at (anchor) {#6};
            \node[xshift=-1.5mm] at (anchor) {$\uparrow$};
      }{}
      \ifthenelse{\isin{#1}{right} \AND \isin{#2}{up}}{
            \coordinate (anchor) at ($({#3})!{#5}!({#4})$);
            \node[darkmsgcircle, xshift=5.5mm] at (anchor) {#6};
            \node[xshift=1.5mm] at (anchor) {$\uparrow$};
      }{}
}

\newcommand{\bwmsg}[6]{
      \ifthenelse{\isin{#1}{left} \AND \isin{#2}{down}}{
            \coordinate (anchor) at ($({#3})!{#5}!({#4})$);
            \node[msgdoublecircle, xshift=-5.5mm] at (anchor) {#6};
            \node[xshift=-1.5mm] at (anchor) {$\downarrow$};
      }{}
      \ifthenelse{\isin{#1}{right} \AND \isin{#2}{down}}{
            \coordinate (anchor) at ($({#3})!{#5}!({#4})$);
            \node[msgdoublecircle, xshift=5.5mm] at (anchor) {#6};
            \node[xshift=1.5mm] at (anchor) {$\downarrow$};
      }{}

      \ifthenelse{\isin{#1}{down} \AND \isin{#2}{right}}{
            \coordinate (anchor) at ($({#3})!{#5}!({#4})$);
            \node[msgdoublecircle, yshift=-6.0mm] at (anchor) {#6};
            \node[yshift=-2.0mm] at (anchor) {$\rightarrow$};
      }{}
      \ifthenelse{\isin{#1}{up} \AND \isin{#2}{right}}{
            \coordinate (anchor) at ($({#3})!{#5}!({#4})$);
            \node[msgdoublecircle, yshift=6.0mm] at (anchor) {#6};
            \node[yshift=2.0mm] at (anchor) {$\rightarrow$};
      }{}

      \ifthenelse{\isin{#1}{down} \AND \isin{#2}{left}}{
            \coordinate (anchor) at ($({#3})!{#5}!({#4})$);
            \node[msgdoublecircle, yshift=-6.0mm] at (anchor) {#6};
            \node[yshift=-2.0mm] at (anchor) {$\leftarrow$};
      }{}
      \ifthenelse{\isin{#1}{up} \AND \isin{#2}{left}}{
            \coordinate (anchor) at ($({#3})!{#5}!({#4})$);
            \node[msgdoublecircle, yshift=6.0mm] at (anchor) {#6};
            \node[yshift=2.0mm] at (anchor) {$\leftarrow$};
      }{}

      \ifthenelse{\isin{#1}{left} \AND \isin{#2}{up}}{
            \coordinate (anchor) at ($({#3})!{#5}!({#4})$);
            \node[msgdoublecircle, xshift=-5.5mm] at (anchor) {#6};
            \node[xshift=-1.5mm] at (anchor) {$\uparrow$};
      }{}
      \ifthenelse{\isin{#1}{right} \AND \isin{#2}{up}}{
            \coordinate (anchor) at ($({#3})!{#5}!({#4})$);
            \node[msgdoublecircle, xshift=5.5mm] at (anchor) {#6};
            \node[xshift=1.5mm] at (anchor) {$\uparrow$};
      }{}
}

\newcommand{\bwdarkmsg}[6]{
      \ifthenelse{\isin{#1}{left} \AND \isin{#2}{down}}{
            \coordinate (anchor) at ($({#3})!{#5}!({#4})$);
            \node[darkmsgdoublecircle, xshift=-5.5mm] at (anchor) {#6};
            \node[xshift=-1.5mm] at (anchor) {$\downarrow$};
      }{}
      \ifthenelse{\isin{#1}{right} \AND \isin{#2}{down}}{
            \coordinate (anchor) at ($({#3})!{#5}!({#4})$);
            \node[darkmsgdoublecircle, xshift=5.5mm] at (anchor) {#6};
            \node[xshift=1.5mm] at (anchor) {$\downarrow$};
      }{}

      \ifthenelse{\isin{#1}{down} \AND \isin{#2}{right}}{
            \coordinate (anchor) at ($({#3})!{#5}!({#4})$);
            \node[darkmsgdoublecircle, yshift=-6.0mm] at (anchor) {#6};
            \node[yshift=-2.0mm] at (anchor) {$\rightarrow$};
      }{}
      \ifthenelse{\isin{#1}{up} \AND \isin{#2}{right}}{
            \coordinate (anchor) at ($({#3})!{#5}!({#4})$);
            \node[darkmsgdoublecircle, yshift=6.0mm] at (anchor) {#6};
            \node[yshift=2.0mm] at (anchor) {$\rightarrow$};
      }{}

      \ifthenelse{\isin{#1}{down} \AND \isin{#2}{left}}{
            \coordinate (anchor) at ($({#3})!{#5}!({#4})$);
            \node[darkmsgdoublecircle, yshift=-6.0mm] at (anchor) {#6};
            \node[yshift=-2.0mm] at (anchor) {$\leftarrow$};
      }{}
      \ifthenelse{\isin{#1}{up} \AND \isin{#2}{left}}{
            \coordinate (anchor) at ($({#3})!{#5}!({#4})$);
            \node[darkmsgdoublecircle, yshift=6.0mm] at (anchor) {#6};
            \node[yshift=2.0mm] at (anchor) {$\leftarrow$};
      }{}

      \ifthenelse{\isin{#1}{left} \AND \isin{#2}{up}}{
            \coordinate (anchor) at ($({#3})!{#5}!({#4})$);
            \node[darkmsgdoublecircle, xshift=-5.5mm] at (anchor) {#6};
            \node[xshift=-1.5mm] at (anchor) {$\uparrow$};
      }{}
      \ifthenelse{\isin{#1}{right} \AND \isin{#2}{up}}{
            \coordinate (anchor) at ($({#3})!{#5}!({#4})$);
            \node[darkmsgdoublecircle, xshift=5.5mm] at (anchor) {#6};
            \node[xshift=1.5mm] at (anchor) {$\uparrow$};
      }{}
}

\allowdisplaybreaks

\title{Active Inference and Epistemic Value in Graphical Models} 

\author[1]{Thijs van de Laar}
\author[1,2]{Magnus Koudahl}
\author[1]{Bart van Erp}
\author[1,3]{Bert de Vries}
\affil[1]{Department of Electrical Engineering, Eindhoven University of Technology, Eindhoven, The Netherlands}
\affil[2]{Nested Mind Solutions, Liverpool, England}
\affil[3]{GN Hearing Benelux BV, Eindhoven, The Netherlands}

\begin{document}

\maketitle

\begin{abstract}
The Free Energy Principle (FEP) postulates that biological agents perceive and interact with their environment in order to minimize a Variational Free Energy (VFE) with respect to a generative model of their environment. The inference of a policy (future control sequence) according to the FEP is known as Active Inference (AIF). The AIF literature describes multiple VFE objectives for policy planning that lead to epistemic (information-seeking) behavior. However, most objectives have limited modeling flexibility. This paper approaches epistemic behavior from a constrained Bethe Free Energy (CBFE) perspective. Crucially, variational optimization of the CBFE can be expressed in terms of message passing on free-form generative models. The key intuition behind the CBFE is that we impose a point-mass constraint on predicted outcomes, which explicitly encodes the assumption that the agent will make observations in the future. We interpret the CBFE objective in terms of its constituent behavioral drives. We then illustrate resulting behavior of the CBFE by planning and interacting with a simulated T-maze environment. Simulations for the T-maze task illustrate how the CBFE agent exhibits an epistemic drive, and actively plans ahead to account for the impact of predicted outcomes. Compared to an EFE agent, the CBFE agent incurs expected reward in significantly more environmental scenarios. We conclude that CBFE optimization by message passing suggests a general mechanism for epistemic-aware AIF in free-form generative models.
\end{abstract}


{\footnotesize \noindent This is the author's version of the article that has been accepted for publication in Frontiers in Robotics and AI.} 

\section{Introduction}
Free energy can be considered as a central concept in the natural sciences. Many natural laws can be derived through the principle of least action\footnote{In this context, ``action'' refers to the path integral, and is distinct from ``action'' in the context of an intervention.}, which rests on variational methods to minimize a path integral of free energy over time \cite{caticha_entropic_2012}. In neuroscience, an application of the least action principle to biological behavior is formalized as the Free Energy Principle \cite{friston_free_2006}. The Free Energy Principle (FEP) postulates that biological agents perceive and interact with their environment in order to minimize a Variational Free Energy (VFE) that is defined with respect to a model of their environment.

Under the FEP, perception relates to the process of hidden state estimation, where the agent tries to infer hidden causes of its sensory observations; and action (intervention) relates to a process where the agent actively tries to influence its (predicted) future observations by manipulating the external environment. Because the future is unobserved (by definition), the agent includes prior beliefs\footnote{We will use ``belief'' and ``distribution'' interchangeably.} about desired outcomes in its model and infers a policy that prescribes a sequence of future controls\footnote{We use ``controls'' refer to quantities in the generative model, and ``actions'' (in the intervention sense) to refer to quantities in the external environment.}. The corollary of the FEP that includes action is referred to as Active Inference (AIF) \cite{friston_action_2010}.

The AIF literature describes multiple Free Energy (FE) objectives for policy planning, e.g., the Expected FE  \cite{friston_active_2015}, Generalized FE \cite{parr_generalised_2019} and Predicted (Bethe) FE \cite{schwobel_active_2018} (among others, see e.g. \cite{tschantz_reinforcement_2020,sajid_bayesian_2021,hafner_action_2020}). Traditionally, the Expected Free Energy (EFE) is evaluated for a selection of policies, and a posterior distribution over policies is constructed from the corresponding EFEs. The EFE is designed to balance epistemic (knowledge seeking) and extrinsic (goal seeking) behavior. The active policy (the sequence of future controls to be executed in the environment) is then selected from this policy posterior \cite{friston_active_2015}.

Several authors have attempted to formulate minimization of the EFE by message passing on factor graphs \cite{de_vries_factor_2017,parr_generalised_2019,parr_neuronal_2019,champion_realising_2021}. These formulations evaluate the EFE objective with the use of a message passing scheme. In this paper we revisit this problem and compare the EFE approach with the message passing interpretation of the variational optimization of a Bethe Free Energy (BFE) \cite{caticha_relative_2004,yedidia_constructing_2005,pearl_probabilistic_1988}. However, the BFE is known to lack epistemic (information-seeking) qualities, and resulting BFE AIF agents therefore do not pro-actively seek informative states \cite{schwobel_active_2018}.

As a solution to the lack of epistemic qualities of the BFE, in this paper we approach epistemic behavior from a \emph{Constrained} BFE (CBFE) perspective \cite{senoz_variational_2021}. We illustrate how optimization of a point-mass constrained BFE objective instigates self-evidencing behavior. Crucially, variational optimization of the CBFE can be expressed in terms of message passing on a graphical representation of the underlying generative model (GM) \cite{dauwels_variational_2007,cox_factor_2019}, without modification of the GM itself. The contributions of this paper are as follows:
\begin{itemize}
\item We formulate the CBFE as an objective for epistemic-aware active inference (Sec.~\ref{sec:cbfe_objective}) that can be interpreted as message passing on a GM (Sec.~\ref{sec:planning_t_maze});
\item We interpret the constituent terms of the CBFE objective as drivers for behavior (Sec.~\ref{sec:epistemic_contribution});
\item We illustrate our interpretation of the CBFE by planning and interacting with a simulated T-maze environment (Sec.~\ref{sec:experimental_setting}).
\item Simulations show that the CBFE agent plans epistemic policies multiple time-steps ahead (Sec.~\ref{sec:inference_results_planning}), and accrues reward for a significantly larger set of scenarios than the EFE (Sec.~\ref{sec:interactive_simulation}).
\end{itemize}

The main advantage of AIF with the CBFE objective, is that it allows inference to be fully automated by message passing, while retaining the epistemic qualities of the EFE. Automated message passing absolves the need for manual derivations and removes computational barriers in scaling AIF to more demanding settings \cite{van_de_laar_simulating_2018}.

\section{Problem Statement}

In this section we will introduce the free energy objectives as used throughout the paper. We start by introducing the Variational Free Energy (VFE), and explain how a VFE can be employed in an AIF context for perception and policy planning. We then introduce the Expected Free Energy (EFE) as a variational objective that is explicitly designed to yield epistemic behavior in AIF agents, but also note that the EFE definition limits itself to (hierarchical) state-space models. We then introduce the Bethe Free Energy (BFE), and argue that the BFE allows for convenient optimization on free-form models by message passing, but note that the BFE lacks information-seeking qualities. We conclude this section by introducing the Constrained Bethe Free Energy (CBFE), which equips the BFE with information-seeking qualities on free-form models through additional constraints on the variational distribution.

Table~\ref{tbl:notation} summarizes notational conventions throughout the paper.
\begin{table}[ht]
\center
\caption{Summary of notational conventions throughout the paper.}
\label{tbl:notation}
\setlength{\extrarowheight}{3pt}
\makebox[\textwidth][c]{
\begin{tabular}{r | l | p{12cm}}
\emph{Notation}                       & \emph{Def.}                                           & \emph{Explanation}\\
\hline
$\bm{s}$                              &                                                       & Collection of (arbitrary) model variables\\
$s_j$                                 &                                                       & Individual model variable with index $j \in \mathcal{S}$\\
$f(\bm{s})$                           & \eqref{eq:general_GM_factorization}                   & Factorized model of variables $\bm{s}$\\
$f_a(\bm{s}_a)$                       & \eqref{eq:general_GM_factorization}                   & Factor (conditional or prior probability distribution) with argument variables $\bm{s}_a$ and index $a \in \mathcal{F}$\\
$q(\bm{s})$                           & \eqref{eq:VFE}                                        & Variational distribution of (latent) variables $\bm{s}$\\
$\U{q(\bm{s})}{f(\bm{s})}$            & \eqref{eq:VFE}                                        & Average energy\\
$\H{q(\bm{s})}$                       & \eqref{eq:VFE}                                        & Entropy\\
$\F[q]$                               & \eqref{eq:VFE}                                        & Variational Free Energy\\
$y_k$, $x_k$, $u_k$                   &                                                       & Observation, state and control variable (at time $k$) respectively\\
$\hat{y}_k$                           &                                                       & Specific realization for observation or unobserved future (predicted) outcome\\
$\hat{u}_k$                           &                                                       & Specific control realization\\
$p(y_k, x_k | x_{k-1}, u_k)$          & \eqref{eq:GM_model_engine}                            & Generative Model engine\\ 
$p(y_k | x_k)$                        & \eqref{eq:GM_model_engine}                            & Observation model\\
$p(x_k | x_{k-1}, u_k)$               & \eqref{eq:GM_model_engine}                            & Transition model\\
$p(x_{t-1})$                          & \eqref{eq:model_perception}                           & State prior\\
$\bm{y}$, $\bm{x}$, $\bm{u}$          &                                                       & Sequence of future observation variables $\bm{y}_{t:t+T-1}$, state variables $\bm{x}_{t-1:t+T-1}$ and control variables $\bm{u}_{t:t+T-1}$, respectively\\
$\hat{\bm{u}}^{j}$                      &                                                       & Policy (sequence of specific future controls), $\hat{\bm{u}}^{j}\in \mathcal{C}$, where index $j$ is usually omitted\\
$\F^{*}(\hat{\bm{u}}^{j})$              & \eqref{eq:F_t_star_bound}                             & Optimized Variational Free Energy for policy $\hat{\bm{u}}^{j}$\\
$\hat{\bm{u}}^*$                      & \eqref{eq:VFE_policy_optim}                           & Optimal policy $\hat{\bm{u}}^* \in \mathcal{C}$\\
$\G[q; \hat{\bm{u}}^{j}]$               & \eqref{eq:EFE_general}                                & Expected Free Energy (EFE)\\
$p(\bm{y} | \bm{x})$                  & \eqref{eq:aggregate_observation_model}                & Aggregate observation model\\
$p(\bm{x} | \bm{u})$                  & \eqref{eq:aggregate_transition_model}                 & Aggregate state transition model, including state prior\\
$\tilde{p}(\bm{y})$                   & \eqref{eq:aggregate_goal_prior}                       & Goal prior for sequence of future observation variables\\
$\tilde{p}(y_k)$                      & \eqref{eq:aggregate_goal_prior}                       & Goal prior for observation variable at time $k$\\
$f(\bm{y}, \bm{x} | \bm{u})$          & \eqref{eq:efe_model_def}, \eqref{eq:generative-model} & Factorized model of future variables (at time $t$), for EFE and (C)BFE respectively\\
$\B[q]$                               & \eqref{eq:BFE_general}                                & Bethe Free Energy (BFE)\\
$\HB{q}$                              & \eqref{eq:HB}                                         & Bethe entropy\\
$\B[q; \hat{\bm{u}}^{j}]$               & \eqref{eq:BFE}                                        & Bethe Free Energy of future model under policy $\hat{\bm{u}}^{j}$\\
$\B[q; \hat{\bm{u}}^{j}, \hat{\bm{y}}]$ & \eqref{eq:CBFE}                                       & Constrained Bethe Free Energy (CBFE) of future model under policy $\hat{\bm{u}}^{j}$ and predicted outcomes $\hat{\bm{y}}$
\end{tabular}}
\end{table}

\subsection{Variational Free Energy}\label{sec:VFE}
The Variational Free Energy (VFE) is a principled metric in physics, where a time-integral over free energy is known as the action functional. Many natural laws can be derived from the principle of least action, where the action functional is minimized with the use of variational calculus \cite{caticha_entropic_2012,lanczos_variational_2012}.

\FloatBarrier
The VFE is defined with respect to a factorized generative model (GM). We consider a GM $f(\bm{s})$ with factors $\{f_a | a\in \mathcal{F}\}$ and variables $\{s_i | i\in \mathcal{S}\}$ that factorizes according to
\begin{align}
    f(\bm{s}) = \prod_{a\in\mathcal{F}} f_a(\bm{s}_a)\,, \label{eq:general_GM_factorization}
\end{align}
where $\bm{s}_a$ collects the argument variables of the factors $f_a$. As a notational convention, we write collections and sequences in bold script. In the model factorization of \eqref{eq:general_GM_factorization}, the factors $f_a$ would correspond with the prior and conditional probability distributions that define the GM. The VFE is then defined as a functional of an (approximate) posterior $q(\bm{s})$ over latent variables, as
\begin{align}
    \F[q] &= \E{q(\bm{s})}{\log \frac{q(\bm{s})}{f(\bm{s})}}\notag\\
    &= \U{q(\bm{s})}{f(\bm{s})} - \H{q(\bm{s})}\,,\label{eq:VFE}
\end{align}
which consists of an average energy $\U{q(\bm{s})}{f(\bm{s})} = -\E{q(\bm{s})}{\log f(\bm{s})}$ and an entropy $\H{q(\bm{s})} = -\E{q(\bm{s})}{\log q(\bm{s})}$.

Because the VFE is (usually) optimized with respect to the posterior $q$ with the use of variational calculus \cite{yedidia_constructing_2005}, the posterior $q$ is also referred to as the \emph{variational distribution}. In this paper, we will strictly reserve the $q$ notation for variational distributions.

We can relate the exact posterior belief with the model definition through a normalizing constant $Z$, as
\begin{align}
    p(\bm{s}) = \frac{f(\bm{s})}{Z}\,, \label{eq:exact_posterior}
\end{align}
where
\begin{align}
    Z = \sum_{\bm{s}} f(\bm{s})\,. \label{eq:evidence}
\end{align}
Throughout this paper, summation can be replaced by integration in the case of continuous variables.

In a Bayesian context, the normalizer $Z$ is commonly referred to as the marginal likelihood or evidence for model $f$. However, exact summation (marginalization) of \eqref{eq:evidence} over all variable realizations is often prohibitively difficult in practice, so that the evidence and exact posterior become unobtainable.

Substituting \eqref{eq:exact_posterior} in the VFE \eqref{eq:VFE} expresses the VFE as an upper bound on the surprise, that is the negative log-model evidence, as
\begin{align}
    \F[q] &= \underbrace{\KL{q(\bm{s})}{p(\bm{s})}}_{\text{posterior divergence}} \underbrace{-\log Z}_{\text{surprise}}\,. \label{eq:VFE_bound}
\end{align}
The marginalization problem of \eqref{eq:evidence} is thus converted to an optimization problem over $q$. After optimization,
\begin{align}
    q^{*} = \arg \min_{q} \F[q]\,, \label{eq:VFE_optim}
\end{align}
the VFE approximates the surprise, and the optimal variational distribution becomes an approximation to the true posterior, $p(\bm{s}) \approx q^{*}(\bm{s})$.

Crucially, we are free to choose constraints on $q$ such that the optimization becomes practically feasible, at the cost of an increased posterior divergence. One such approximation is the Bethe assumption, as we will see in Sec.~\ref{sec:BFE}.

\subsection{Inference for Perception}
\label{sec:inference_perception}
We now return to the model definition of \eqref{eq:general_GM_factorization}. In the context of AIF, a Generative Model (GM) comprises of a probability distribution over states $x_k$, observations $y_k$ and controls $u_k$, at each time index $k$. We will use a hat to indicate specific variable realizations, i.e. $\hat{y}_k$ for a specific outcome and $\hat{u}_k$ for a specific control at time $k$. As a notational convention, we use $k$ as an arbitrary time index, often used in the context of iterations, and we use $t$ to indicate the current simulation time index.

We define a state-space model \cite{koller_probabilistic_2009} for the generative model engine, which represents our belief about how observations follow from a given control and previous state, as
\begin{align}
    p(y_k, x_k | x_{k-1}, u_k) = \underbrace{p(y_k | x_k)}_{\substack{\text{observation}\\\text{model}}}\, \underbrace{p(x_k | x_{k-1}, u_k)}_{\substack{\text{transition}\\\text{model}}}\,. \label{eq:GM_model_engine}
\end{align}

We use a prior belief about past states $p(x_{t-1})$ together with the generative model engine \eqref{eq:GM_model_engine} to define a generative model for perception
\begin{align}
    f(y_t, x_t, x_{t-1} | u_t) &= p(x_{t-1}) \,p(y_t, x_t | x_{t-1}, u_t)\notag \\
    &= p(x_{t-1}) \, p(y_t | x_t)\, p(x_t | x_{t-1}, u_t)\,, \label{eq:model_perception}
\end{align}
which, after substitution in \eqref{eq:VFE}, results in the VFE objective for perception,
\begin{align}
    \F[q] &= \U{q(x_t, x_{t-1})}{f(\hat{y}_t, x_t, x_{t-1} | \hat{u}_t)} - \H{q(x_t, x_{t-1})}\,. \label{eq:VFE_perception}
\end{align}

At each time $t$, the process of perception then relates to inferring the optimal variational distribution $q^{*}(x_t, x_{t-1})$ about latent states, given the current action $\hat{u}_t$ and resulting outcome $\hat{y}_t$. The resulting variational distribution can then be used as a prior for the next time-step, such that $p(x_t) \triangleq q^{*}(x_t) = \sum_{x_{t-1}} q^{*}(x_t, x_{t-1})$\,.

\subsection{Inference for Planning}
\label{sec:inference_planning}
At each time $t$, \emph{planning} is concerned with selecting optimal future controls by minimizing a Free Energy (FE) objective that is defined with respect to future variables. We write $\bm{y} = \bm{y}_{t:t+T-1}$, $\bm{x} = \bm{x}_{t-1:t+T-1}$, and $\bm{u} = \bm{u}_{t:t+T-1}$ as the sequences of future observations, states and controls respectively, for a fixed-time horizon of $T$ time-steps ahead.

We will refer to a specific future control sequence $\hat{\bm{u}}$ as a \emph{policy}. The optimal policy $\hat{\bm{u}}^*$ is then referred to as the \emph{active} policy, where (local) optimality is indicated by an asterisk. Inference for planning then aims to select the optimal policy (in terms of FE) from a collection of candidate policies $\hat{\bm{u}}^{j}\in\mathcal{C}$, where $\mathcal{C}$ represents the finite set of all (user-provided) candidate policies, and $j$ the selected policy index.

When we view the candidate policy $\hat{\bm{u}}^{j}$ as a model selection variable, the problem of policy selection becomes equivalent to the problem of Bayesian model selection, where we wish to find a probabilistic model with the highest posterior probability among some given candidate models. When there is no prior preference about models, the optimal model is the one with the highest marginal likelihood (evidence).

Given a model $f(\bm{y}, \bm{x} | \bm{u})$ of future observations and states given a future control sequence, we can express the marginal likelihood (evidence) for a specific policy choice, as
\begin{align}
    Z_j = \sum_{\bm{y}} \sum_{\bm{x}} f(\bm{y}, \bm{x} | \hat{\bm{u}}^{j})\,.
\end{align}
Using \eqref{eq:exact_posterior}, we can then relate the exact posterior belief with the variational distribution and the policy evidence, as
\begin{align}
    p(\bm{y}, \bm{x} | \hat{\bm{u}}^{j}) = \frac{f(\bm{y}, \bm{x} | \hat{\bm{u}}^{j})}{Z_j}\,.
\end{align}
Under optimization of $q$, the minimal VFE then approximates the surprise \eqref{eq:VFE_bound}, as
\begin{align}
    \F^{*}(\hat{\bm{u}}^{j}) = \min_{q} \F[q; \hat{\bm{u}}^{j}] \geq -\log Z_j\,. \label{eq:F_t_star_bound}
\end{align}
The optimal policy then minimizes the optimized VFE, as
\begin{align}
    \hat{\bm{u}}^* = \argmin_{\hat{\bm{u}}^{j}\in\mathcal{C}}\F^{*}(\hat{\bm{u}}^{j}) \,. \label{eq:VFE_policy_optim}
\end{align}
In the following, we omit the explicit indexing of the policy on $j$ for notational convenience, and simply write $\hat{\bm{u}}$ to represent a specific policy choice.

Because the VFE \eqref{eq:VFE} involves expectations over the full joint variational distribution, it may become prohibitively expensive to compute for larger models. Therefore, additional assumptions and constraints on the VFE are often required. As a result, multiple free energy objectives for policy planning have been proposed in the literature, e.g., the Expected Free Energy (EFE) \cite{friston_active_2015,friston_sophisticated_2021}, the Free Energy of the Expected Future \cite{millidge_whence_2020}, the Generalized Free Energy \cite{parr_generalised_2019}, the Predicted (Bethe) Free Energy \cite{schwobel_active_2018}, and marginal approximations \cite{parr_neuronal_2019}.

\subsection{Expected Free Energy}
\label{sec:efe_objective}
The Expected Free Energy (EFE) is an FE objective for planning that is explicitly constructed to elicit information-seeking behavior \cite{friston_active_2015}. Because future observations are (by definition) unknown, the EFE is defined in terms of an expectation that includes observation variables, as
\begin{align}
    \G[q; \hat{\bm{u}}] = \E{q(\bm{y}, \bm{x})}{\log \frac{q(\bm{x})}{f(\bm{y}, \bm{x} | \hat{\bm{u}})}}\,. \label{eq:EFE_general}
\end{align}

Construction of the (Markovian) model for the EFE starts by stringing together a state prior with the generative model engine of \eqref{eq:GM_model_engine} for future times, as
\begin{align}
    p(\bm{y}, \bm{x} | \bm{u}) &= p(x_{t-1}) \prod_{k=t}^{t+T-1} p(y_k, x_k| x_{k-1}, u_k)\notag\\
    &= p(x_{t-1}) \prod_{k=t}^{t+T-1} p(y_k | x_k)\, p(x_k | x_{k-1}, u_k)\,,\label{eq:stringed_gm_engine}
\end{align}
where the state prior $p(x_{t-1})$ follows from the perceptual process (Sec.~\ref{sec:inference_perception}). For notational convenience, we often group the observation and state transition models (including the state prior), according to
\begin{subequations}
\begin{align}
    p(\bm{y} | \bm{x}) &= \prod_{k=t}^{t+T-1} p(y_k | x_k) \label{eq:aggregate_observation_model}\\
    p(\bm{x} | \bm{u}) &= p(x_{t-1}) \prod_{k=t}^{t+T-1} p(x_k | x_{k-1}, u_k) \label{eq:aggregate_transition_model}\,.
\end{align}
\end{subequations}

From the future generative model engine \eqref{eq:stringed_gm_engine}, the EFE defines a state posterior
\begin{align}
    p(\bm{x} | \bm{y}, \bm{u}) &= \frac{p(\bm{y}, \bm{x} | \bm{u})}{\sum_{\bm{x}} p(\bm{y}, \bm{x} | \bm{u})} \label{eq:post_pred_a}\,. 
\end{align}
Note that our notation differs from \cite{friston_active_2015}, where posterior distributions are denoted by $q$. We strictly reserve the $q$ notation for variational distributions.

We introduce goal priors $\tilde{p}(y_k)$ over observation variables. Goal priors encode prior beliefs about desired observations (or states) \cite{friston_anatomy_2013}, and are annotated with a tilde to avoid confusion with the marginal distribution over the same variable. We then introduce a shorthand notation that aggregates (independent) goal priors for the future generative model, as
\begin{align}
    \tilde{p}(\bm{y}) &= \prod_{k=t}^{t+T-1} \tilde{p}(y_k)\,. \label{eq:aggregate_goal_prior}
\end{align}
Together with the aggregated goal prior, the factorized model for the EFE is then constructed as
\begin{align}
    f(\bm{y}, \bm{x} | \bm{u}) &= p(\bm{x} | \bm{y}, \bm{u})\, \tilde{p}(\bm{y})\,. \label{eq:efe_model_def}
\end{align}
There are several things of note about the model of \eqref{eq:efe_model_def}:
\begin{itemize}
    \item The model includes variables that pertain to future time-points, $t \leq k \leq t+T-1$. As a result, the future observation variables $\bm{y}$ are latent;
    \item The model includes a state prior that is a result of inference for perception;
    \item The (informative) goal priors $\tilde{p}$ introduce a bias in the model towards desired outcomes;
    \item Candidate policies will be given, as indicated by a conditioning on controls.
\end{itemize}

Upon substitution of \eqref{eq:efe_model_def}, the EFE \eqref{eq:EFE_general} factorizes into an epistemic and an extrinsic value term \cite{friston_active_2015}, as
\begin{align}
    \G[q; \hat{\bm{u}}] = -\underbrace{\E{q(\bm{y}, \bm{x})}{\log \frac{p(\bm{x} | \bm{y}, \hat{\bm{u}})}{q(\bm{x})}}}_{\text{epistemic value}} - \underbrace{\E{q(\bm{y})}{\log \tilde{p}(\bm{y})}}_{\text{extrinsic value}}\,, \label{eq:EFE_ep_ex}
\end{align}
where the epistemic value relates to a mutual information between states and observations. This decomposition is often used to motivate the epistemic qualities of the EFE.

An alternative decomposition, in terms of ambiguity and observation risk, can be obtained under the assumptions $q(\bm{y} | \bm{x}) \approx p(\bm{y} | \bm{x})$ (approximation of the observation model), and $q(\bm{x} | \bm{y}) \approx p(\bm{x} | \bm{y}, \bm{u})$ (approximation of the exact posterior). These assumptions allow us to rewrite the exact relationship $q(\bm{y}, \bm{x}) = q(\bm{y} | \bm{x})\, q(\bm{x}) = q(\bm{x} | \bm{y})\, q(\bm{y})$ in terms of the approximations $q(\bm{y}, \bm{x}) \approx p(\bm{y} | \bm{x})\, q(\bm{x}) \approx p(\bm{x} | \bm{y}, \bm{u})\, q(\bm{y})$. As a result, we obtain
\begin{align}
    \G[q; \hat{\bm{u}}] &\approx \E{q(\bm{y}, \bm{x})}{\log \frac{q(\bm{y})}{p(\bm{y} | \bm{x})\, \tilde{p}(\bm{y})}}\notag\\
    &\approx \underbrace{\E{q(\bm{x})}{\H{p(\bm{y}| \bm{x})}}}_{\text{ambiguity}} + \underbrace{\KL{q(\bm{y})}{\tilde{p}(\bm{y})}}_{\text{observation risk}}\,, \label{eq:EFE_amb_risk}
\end{align}
where $q(\bm{x})$ and $q(\bm{y})$ on the r.h.s. are implicitly conditioned on $\hat{\bm{u}}$. This decomposition is often used to motivate the explorative (ambiguity reducing) qualities of the EFE.

In the current paper we evaluate the EFE in accordance with \cite{friston_active_2015,da_costa_active_2020}, for which the procedure is detailed in Appendix~A.

Although the EFE leads to epistemic behavior, it does not fit the general functional form of the VFE \eqref{eq:VFE}, where the expectation and numerator define the same variational distribution. As a result, EFE minimization by message passing requires custom definitions, and limits itself to (hierarchical) state-space models. Furthermore, note that the EFE involves the state posterior $p(\bm{x} | \bm{y}, \bm{u})$ as part of its definition, which is technically a quantity that needs to be inferred. The EFE thus conflates the definition of the planning objective with the inference procedure for planning itself.

\subsection{Bethe Free Energy}\label{sec:BFE}
The Bethe Free Energy (BFE) defines a variational distribution that factorizes according to the Bethe assumption
\begin{align}
    q(\bm{s}) \triangleq \prod_{a\in \mathcal{F}} q_a(\bm{s}_a)\prod_{i\in \mathcal{S}} q_i(s_i)^{1-d_i}\,, \label{eq:bethe_assumption}
\end{align}
where the degree $d_i$ counts how many $q_a$'s contain $s_i$ as an argument. After substituting the Bethe assumption \eqref{eq:bethe_assumption} in the VFE \eqref{eq:VFE}, we obtain the BFE,
\begin{align}
    \B[q] &= \sum_{a\in \mathcal{F}}\U{q_a(\bm{s}_a)}{f_a(\bm{s}_a)} - \sum_{a\in \mathcal{F}} \H{q_a(\bm{s}_a)} - \sum_{i\in \mathcal{S}} (1 - d_i) \H{q_i(s_i)}\,, \label{eq:BFE_general}
\end{align}
as a special case of the VFE. The entropy contributions are often summarized in the Bethe entropy, as
\begin{align}
    \HB{q} = \sum_{a\in \mathcal{F}} \H{q_a(\bm{s}_a)} + \sum_{i\in \mathcal{S}} (1 - d_i) \H{q_i(s_i)}\,. \label{eq:HB}
\end{align}

Because the BFE fully factorizes into local contributions in $\mathcal{F}$ and $\mathcal{S}$, it can be optimized by message passing on the generative model \cite{pearl_probabilistic_1988,yedidia_constructing_2005,wainwright_graphical_2008,senoz_variational_2021}. In the context of AIF, the BFE for a model over future states is also referred to as the Predicted Free Energy \cite{schwobel_active_2018}.

For a fixed time-horizon $T$, the factorized model for future states is constructed from the generative model engine and goal prior, as
\begin{align}
    f(\bm{y}, \bm{x} | \bm{u}) &= p(\bm{y}, \bm{x} | \bm{u})\, \tilde{p}(\bm{y}) \notag\\
    &= p(x_{t-1}) \prod_{k=t}^{t+T-1} p(y_k, x_k | x_{k-1}, u_k)\, \tilde{p}(y_k) \notag\\
    &= p(x_{t-1}) \prod_{k=t}^{t+T-1} p(y_k | x_k)\, p(x_k | x_{k-1}, u_k)\, \tilde{p}(y_k)\,. \label{eq:generative-model}
\end{align}
Because the generative model engine and goal priors introduce a simultaneous constraint on future observations, the model of \eqref{eq:generative-model} represents a scaled probability distribution. The BFE of the future model under policy $\hat{\bm{u}}$ then becomes
\begin{align}
    \B[q; \hat{\bm{u}}] = \U{q(\bm{y}, \bm{x})}{f(\bm{y}, \bm{x} | \bm{u})} - \HB{q(\bm{y}, \bm{x})}\,. \label{eq:BFE}
\end{align}

A major advantage of the BFE over the EFE as an objective for AIF is that message passing implementations can be automatically derived on free form models, thus greatly enhancing model flexibility. A drawback of the BFE, however, is that it lacks the epistemic qualities of the EFE \cite{schwobel_active_2018}, see also Sec.~\ref{sec:epistemic_contribution}.

\subsection{Constrained Bethe Free Energy}\label{sec:cbfe_objective}
The \emph{Constrained} Bethe Free Energy (CBFE) that we propose in this paper combines the epistemic qualities of the EFE with the computational ease and model flexibility of the BFE. The CBFE can be derived from the BFE by imposing additional constraints on the variational distribution
\begin{align}
    q(\bm{y}, \bm{x}) &\triangleq q(\bm{x})\, \delta(\bm{y} - \hat{\bm{y}})\notag\\
    &= q(\bm{x})\! \prod_{k=t}^{t+T-1} \!\delta(y_k - \hat{y}_k)\,, \label{eq:q_con_fact}
\end{align}
where $\delta(\cdot)$ defines the appropriate (Kronecker or Dirac) delta function for the domain of the observation variable $y_k$ (discrete or continuous). The point-mass (delta) constraints of the CBFE are motivated by the following key insight: although the future is unknown, we know that we will observe something in the future. However, because future outcomes are by definition unobserved, the $\hat{y}_k$ encode potential outcomes that need to be optimized for.

For the model of \eqref{eq:generative-model}, the CBFE then becomes\footnote{\label{ftn:entropy}For continuous variables we need to additionally assume that the entropy of a Dirac delta $\H{\delta(\cdot)} = 0$ \cite{senoz_variational_2021}.}
\begin{align}
    \B[q; \hat{\bm{u}}, \hat{\bm{y}}] &= \U{q(\bm{x})\, \delta(\bm{y} - \hat{\bm{y}})}{f(\bm{y}, \bm{x} | \hat{\bm{u}})} - \HB{q(\bm{x})\,\delta(\bm{y} - \hat{\bm{y}})}\notag \\
    &= \U{q(\bm{x})}{f(\hat{\bm{y}}, \bm{x} | \hat{\bm{u}})} - \HB{q(\bm{x})}\,. \label{eq:CBFE}
\end{align}

The current paper investigates how point-mass constraints of the form \eqref{eq:q_con_fact} affect epistemic behavior in AIF agents.

\section{Methods}
\label{sec:methods}
To minimize the (Constrained) Bethe Free Energy, the current paper uses message passing on a Forney-style factor graph (FFG) representation \cite{forney_codes_2001} of the factorized model \eqref{eq:general_GM_factorization}. In an FFG, edges represent variables and nodes represent the functional relationships between variables (i.e. the prior and conditional probabilities).

Especially in a signal processing and control context, the FFG paradigm leads to convenient message passing formulations \cite{korl_factor_2005,loeliger_factor_2007-1}. Namely, inference can be described in terms of messages that summarize and propagate information across the FFG. The BFE is well-known for being the fundamental objective of the celebrated sum-product algorithm \cite{pearl_probabilistic_1988}, which has been formulated in terms of message passing on FFGs \cite{loeliger_introduction_2004}. Extensions of the sum-product algorithm to hybrid formulations, such as variational message passing (VMP) \cite{dauwels_variational_2007} and expectation maximization (EM) \cite{dauwels_expectation_2005} have also been formulated as message passing on FFGs. More recently, more general hybrid algorithms have been described in terms of message passing, see e.g. \cite{zhang_unifying_2017,van_de_laar_chance-constrained_2021}. A comprehensive overview is provided in \cite{senoz_variational_2021}, where additional constraints, including point-mass constraints, are imposed on the BFE and optimized for by message passing on FFGs.

\subsection{Forney-Style Factor Graph Example}
Let us consider an example model \eqref{eq:general_GM_factorization} that factorizes according to
\begin{align}
    f(s_1, s_2, s_3, s_4) = f_a(s_1)\, f_b(s_1, s_2, s_3)\, f_c(s_3)\, f_d(s_2, s_4)\,. \label{eq:FFG_example_fact}
\end{align}
The FFG representation of \eqref{eq:FFG_example_fact} is depicted in Fig.~\ref{fig:FFG_example} (left).
\begin{figure}[ht]
    \hfill
    \makebox[\textwidth][c]{
    \begin{minipage}[b]{0.45\textwidth}
        \centering
        \begin{tikzpicture}
            [node distance=20mm,auto]

            \node[box] (fa) {$f_a$};
            \node[box, right of=fa] (fb) {$f_b$};
            \node[box, below of=fb] (fc) {$f_c$};
            \node[box, right of=fb] (fd) {$f_d$};
            \node[below of=fd] (fe) {};

            \draw[line] (fa) -- node[anchor=south]{$s_1$}(fb);
            \draw[line] (fb) -- node[anchor=west]{$s_3$}(fc);
            \draw[line] (fb) -- node[anchor=south]{$s_2$}(fd);
            \draw[line] (fd) -- node[anchor=west]{$s_4$}(fe);
        \end{tikzpicture}
    \end{minipage}
    \begin{minipage}[b]{0.45\textwidth}
        \centering
        \begin{tikzpicture}
            [node distance=20mm,auto]

            \node[box] (fa) {$f_a$};
            \node[box, right of=fa] (fb) {$f_b$};
            \node[box, below of=fb] (fc) {$f_c$};
            \node[box, right of=fb] (fd) {$f_d$};
            \node[clamped, below of=fd] (fe) {};

            \draw[line] (fa) -- node[anchor=south, pos=0.7]{$s_1$}(fb);
            \draw[line] (fb) -- node[anchor=east, pos=0.75]{$s_3$}(fc);
            \draw[line] (fb) -- node[anchor=south, pos=0.3]{$s_2$}(fd);
            \draw[line] (fd) -- node[anchor=east, pos=0.6]{$s_4$}(fe);

            \node[optim, below of=fb, node distance=10mm] (optim) {$\delta$};

            \msg{up}{right}{fa}{fb}{0.35}{1}
            \darkmsg{up}{right}{fb}{fd}{0.65}{2}
            \msg{right}{up}{fd}{fe}{0.5}{3}
            \msg{down}{left}{fb}{fd}{0.65}{4}
            \darkmsg{down}{left}{fa}{fb}{0.35}{5}
            \darkmsg{left}{down}{fb}{fc}{0.33}{6}
            \msg{right}{up}{fb}{fc}{0.67}{7}
        \end{tikzpicture}
    \end{minipage}}
    \caption{Forney-style factor graph representation for the example model of \eqref{eq:FFG_example_fact} (left) and message passing schedule for the Bethe Free Energy minimization of \eqref{eq:BFE_example_fact} (right). Shaded messages indicate variational message updates, and the solid square node indicates given (clamped) values. The round node indicates a point-mass constraint for which the value is optimized.}
    \label{fig:FFG_example}
\end{figure}

Now suppose we observe $s_4$ and introduce a point-mass constraint on $s_3$. The variational distribution then factorizes as
\begin{align}
    q(s_1, s_2, s_3) &= q(s_1, s_2)\, q(s_3)\notag\\
    &= q(s_1, s_2)\, \delta(s_3 - \hat{s}_3)\,, \label{eq:BFE_example_fact}
\end{align}
where $s_4$ is excluded from the variational distribution because it is observed and therefore no longer a latent variable. Substituting \eqref{eq:FFG_example_fact} and \eqref{eq:BFE_example_fact} in \eqref{eq:CBFE} yields the CBFE as$^{\text{\ref{ftn:entropy}}}$
\begin{align}
    \B[q; \hat{s}_3] &= \U{q(s_1, s_2)}{f(s_1, s_2, \hat{s}_3, \hat{s}_4)} - \H{q(s_1, s_2)}\,,\label{eq:BFE_example_specific}
\end{align}
where we directly substituted the observed value $\hat{s}_4$ into the factorized model. 

In this paper we adhere to the notation in \cite{senoz_variational_2021}, and indicate point-mass constraints by an unshaded round node with an annotated $\delta$ on the corresponding edge of the FFG. A solid square node indicates a given value (e.g., an action, observed outcome or given parameter), whereas an unshaded round node indicates a point-mass constraint that is optimized for (e.g. a potential outcome). Unshaded messages indicate sum-product messages \cite{loeliger_introduction_2004} and shaded messages indicate variational messages, as scheduled and computed in accordance with \cite{dauwels_variational_2007}. The ForneyLab probabilistic programming toolbox \cite{cox_factor_2019} implements an automated message passing scheduler and a lookup table of pre-derived message updates \cite[App.~A]{korl_factor_2005,van_de_laar_automated_2019}.

Variational optimization of \eqref{eq:BFE_example_specific} then yields the (iterative) message passing schedule of Fig.~\ref{fig:FFG_example} (right), where $\hat{s}_3$ is initialized. After computation of the messages, the mode of the product between message $\smalldarkcircled{6}$ and $\smallcircled{7}$ becomes the new value for $\hat{s}_3$, and the schedule is repeated until convergence. The resulting optimization procedure then resembles an expectation maximization (EM) scheme where $\smalldarkcircled{6}$ acts as a likelihood and $\smallcircled{7}$ as a prior \cite{dauwels_expectation_2005}, and where upon each iteration the value $\hat{s}_3$ is updated with the maximum a-posteriori (MAP) estimate.

\section{Value Decompositions}
\label{sec:epistemic_contribution}

In this section we further investigate the drivers for behavior of the (C)BFE. We assume that all variational distributions factorize according to the Bethe assumption \eqref{eq:bethe_assumption}.

\subsection{Confidence and Complexity}
We substitute the model of \eqref{eq:generative-model} in the CBFE definition of \eqref{eq:CBFE} and combine to identify three terms, as
\begin{align}
    \B[q; \hat{\bm{y}}, \hat{\bm{u}}] &= \U{q(\bm{x})\,\delta(\bm{y} - \hat{\bm{y}})}{p(\bm{y}, \bm{x} | \hat{\bm{u}})} + \U{\delta(\bm{y} - \hat{\bm{y}})}{\tilde{p}(\bm{y})} - \HB{q(\bm{x})\delta(\bm{y} - \hat{\bm{y}})}\notag\\
    &= \U{q(\bm{x})\, \delta(\bm{y} - \hat{\bm{y}})}{p(\bm{y}| \bm{x})} - \HB{\delta(\bm{y} - \hat{\bm{y}})} + \U{q(\bm{x})}{p(\bm{x}| \hat{\bm{u}})} - \HB{q(\bm{x})} + \U{\delta(\bm{y} - \hat{\bm{y}})}{\tilde{p}(\bm{y})}\notag\\
    &= \underbrace{\E{q(\bm{x})}{\KL{\delta(\bm{y} - \hat{\bm{y}})}{p(\bm{y} | \bm{x})}}}_{\text{negative confidence}} + \underbrace{\KL{q(\bm{x})}{p(\bm{x} | \hat{\bm{u}})}}_{\text{complexity}} - \underbrace{\log \tilde{p}(\hat{\bm{y}})}_{\text{extrinsic value}}\,. \label{eq:CBFE_confidence_complexity}
\end{align}

Table \ref{tbl:CBFE} is provided as an overview, and summarizes the properties of the individual terms of \eqref{eq:CBFE_confidence_complexity} under optimization.

The extrinsic value induces a preference for extrinsically rewarding future outcomes.

Minimizing complexity prefers policies that induce transitions that are in line with state beliefs, and (vice versa) prefers state beliefs that remain close the policy-induced state transitions.

The confidence expresses the expected difference (divergence) in information between the outcomes as predicted by the observation model and the most likely (expected) outcome. In other words, this term quantifies the information difference between predictions and absolute certainty about outcomes. While the negative confidence term could be interpreted as an ambiguity (deviation from certainty), we choose this alternative terminology to prevent confusion with the ambiguity as defined in \eqref{eq:EFE_amb_risk}.

Specifically, the ambiguity \eqref{eq:EFE_amb_risk} and negative confidence \eqref{eq:CBFE_confidence_complexity} are both of the form$^{\text{\ref{ftn:entropy}}}$ $\U{q(\bm{y}, \bm{x})}{p(\bm{y} | \bm{x})} = -\E{q(\bm{y}, \bm{x})}{\log p(\bm{y} | \bm{x})}$. However, where the ambiguity approximates $q(\bm{y}, \bm{x}) \approx p(\bm{y} | \bm{x})\, q(\bm{x})$ \eqref{eq:EFE_amb_risk}, the confidence defines $q(\bm{y}, \bm{x}) \triangleq q(\bm{x})\, \delta(\bm{y} - \hat{\bm{y}})$ \eqref{eq:q_con_fact}. As a result, the ambiguity explicitly accounts for the full spread of $p(\bm{y} | \bm{x})$, whereas the confidence evaluates $p(\bm{y} | \bm{x})$ at the expected maximum $\hat{\bm{y}}$.

Maximizing confidence prefers outcomes that are in line with predictions, and simultaneously tries to maximize the precision of state beliefs (Table \ref{tbl:CBFE}), see also \cite[p.~2093]{friston_post_2011}. Note that all terms act in unison -- the precision of state beliefs is simultaneously influenced by the complexity, which prevents the collapse of the state belief to a point-mass.

\begin{table}[ht]
\center
\caption{Optima for the individual terms of the CBFE decompositions \eqref{eq:CBFE_confidence_complexity}, \eqref{eq:CBFE_in_ex}. Each row varies one quantity (variable or function) in their respective term while other quantities remain fixed. Shaded cells indicate that the term (row) is not a function(al) of that specific optimization quantity (column).}
\label{tbl:CBFE}
\setlength{\extrarowheight}{3pt}
\makebox[\textwidth][c]{
\begin{tabular}{p{2cm} | p{1.3cm} | p{4cm} | p{4cm} | p{4cm}}
                                                               &                                & \multicolumn{3}{l}{\emph{Vary}}\\
\emph{Optimize}                                                & \emph{Fix}                     & $\hat{\bm{y}}$                                                                                                              & $\hat{\bm{u}}$                                                                                                   & $q(\bm{x})$ \\
\hline\hline
max ex. val.                                                   & \notableentry                  & $\hat{\bm{y}}$ that maximizes $\tilde{p}(\hat{\bm{y}})$                                                                     & \notableentry                                                                                                    & \notableentry \\
\hline\hline
\multirow{3}{*}{\shortstack{min\\complexity}}              & $q(\bm{x})$                    & \notableentry                                                                                                               & $\hat{\bm{u}} \in \mathcal{C}$ that renders $p(\bm{x} | \hat{\bm{u}})$ closest to $q(\bm{x})$                    & \\
\cline{2-5}
                                                               & $\hat{\bm{u}}$                 & \notableentry                                                                                                               &                                                                                                                  & $q(\bm{x}) = p(\bm{x} | \hat{\bm{u}})$ \\
\hline\hline
\multirow{6}{*}{\shortstack{max\\confidence}}              & $q(\bm{x})$                    & $\hat{\bm{y}}$ that maximizes expected outcomes$^{\text{\ref{ftn:entropy}}}$ $\E{q(\bm{x})}{\log p(\hat{\bm{y}} | \bm{x})}$ & \notableentry                                                                                                    & \\
\cline{2-5}
                                                               & $\hat{\bm{y}}$                 &                                                                                                                             & \notableentry                                                                                                    & $q(\bm{x}) = \delta(\bm{x} - \hat{\bm{x}})$ where $\hat{\bm{x}}$ maximizes the likelihood $p(\hat{\bm{y}} | \bm{x})$ \\
\hline\hline
\multirow{5}{*}{\shortstack{max\\intrinsic\\value}} & $\hat{\bm{u}}$                 & $\hat{\bm{y}}$ that maximizes the evidence$^{\text{\ref{ftn:entropy}}}$ $p(\hat{\bm{y}}| \hat{\bm{u}})$                     &                                                                                                                  & \notableentry \\
\cline{2-5}
                                                               & $\hat{\bm{y}}$                 &                                                                                                                             & $\hat{\bm{u}} \in \mathcal{C}$ that renders $p(\bm{y}| \hat{\bm{u}})$ closest to $\delta(\bm{y} - \hat{\bm{y}})$ & \notableentry \\
\hline\hline
\multirow{6}{*}{\shortstack{min\\posterior\\divergence}}       & $q(\bm{x})$, $\hat{\bm{u}}$    & $\hat{\bm{y}}$ that renders $p(\bm{x} | \hat{\bm{y}}, \hat{\bm{u}})$ closest to $q(\bm{x})$                                 &                                                                                                                  & \\
\cline{2-5}
                                                               & $q(\bm{x})$, $\hat{\bm{y}}$    &                                                                                                                             & $\hat{\bm{u}} \in \mathcal{C}$ that renders $p(\bm{x} | \hat{\bm{y}}, \hat{\bm{u}})$ closest to $q(\bm{x})$      & \\
\cline{2-5}
                                                               & $\hat{\bm{y}}$, $\hat{\bm{u}}$ &                                                                                                                             &                                                                                                                  & $q(\bm{x}) = p(\bm{x} | \hat{\bm{y}}, \hat{\bm{u}})$
\end{tabular}}
\end{table}

\FloatBarrier
\subsection{Intrinsic and Extrinsic Value}
\label{sec:in_ex_val}

A second decomposition of the CBFE objective follows when we rewrite the factorized model of \eqref{eq:generative-model} using the product rule, as
\begin{align}
    f(\bm{y}, \bm{x} | \bm{u}) &= p(\bm{y}, \bm{x}| \bm{u})\, \tilde{p}(\bm{y})\notag\\
    &= p(\bm{x} | \bm{y}, \bm{u})\, p(\bm{y} | \bm{u})\, \tilde{p}(\bm{y})\,. \label{eq:gm_posterior}
\end{align}

Substituting \eqref{eq:gm_posterior} in the CBFE definition \eqref{eq:CBFE} and combining terms, then yields
\begin{align}
    \B[q; \hat{\bm{y}}, \hat{\bm{u}}] &= \U{q(\bm{x})\,\delta(\bm{y} - \hat{\bm{y}})}{p(\bm{y}, \bm{x} | \hat{\bm{u}})} + \U{\delta(\bm{y} - \hat{\bm{y}})}{\tilde{p}(\bm{y})} - \HB{q(\bm{x})\delta(\bm{y} - \hat{\bm{y}})}\notag\\
    &= \U{q(\bm{x})}{p(\bm{x} | \hat{\bm{y}}, \hat{\bm{u}})} - \HB{q(\bm{x})} + \U{\delta(\bm{y} - \hat{\bm{y}})}{p(\bm{y} | \hat{\bm{u}})} - \HB{\delta(\bm{y} - \hat{\bm{y}})} + \U{\delta(\bm{y} - \hat{\bm{y}})}{\tilde{p}(\bm{y})}\notag\\
    &= \underbrace{\KL{q(\bm{x})}{p(\bm{x}| \hat{\bm{y}}, \hat{\bm{u}})}}_{\text{posterior divergence}} + \underbrace{\KL{\delta(\bm{y} - \hat{\bm{y}})}{p(\bm{y}| \hat{\bm{u}})}}_{\text{negative intrinsic value}} - \underbrace{\log \tilde{p}(\hat{\bm{y}})}_{\text{extrinsic value}}\,. \label{eq:CBFE_in_ex}
\end{align}

Table \ref{tbl:CBFE} again summarizes the properties of the individual terms of \eqref{eq:CBFE_in_ex} under optimization.

The second term of \eqref{eq:CBFE_in_ex} expresses the difference (divergence) in information between the predicted outcomes and the point-mass constrained (expected) outcome. In contrast to the extrinsic value (third term), this term quantifies a (negative) intrinsic value that purely depends upon the agent's intrinsic beliefs about the environment (state prior and generative model engine \eqref{eq:stringed_gm_engine}). Under optimization (Table \ref{tbl:CBFE}), this term prefers policies that lead to precise predictions for the outcomes.

The posterior divergence (first term) is always non-negative and will diminish under optimization, which allows us to combine \eqref{eq:CBFE_confidence_complexity} and \eqref{eq:CBFE_in_ex} into$^{\text{\ref{ftn:entropy}}}$
\begin{align}
    \underbrace{\log p(\hat{\bm{y}}| \hat{\bm{u}})}_{\substack{\text{predicted}\\\text{log-evidence}}} &\geq \underbrace{\E{q(\bm{x})}{\log p(\hat{\bm{y}} | \bm{x})}}_{\text{confidence}} - \underbrace{\KL{q(\bm{x})}{p(\bm{x} | \hat{\bm{u}})}}_{\text{complexity}}\,. \label{eq:epistemic_balance}
\end{align}
Interestingly, \eqref{eq:epistemic_balance} tells us that the intrinsic value of \eqref{eq:CBFE_in_ex} relates to the model evidence as predicted under the policy and resulting expected outcomes. Inference for planning with the CBFE then attempts to make precise predictions for outcomes by maximizing predicted model evidence. In this view, the CBFE for planning can be considered (quite literally) as self-evidencing \cite{hohwy_self-evidencing_2016,friston_action_2010}. As a result of selected actions, environmental outcomes may still be surprising under current generative model assumptions. Inference for perception then subsequently corrects the generative model priors (Sec.~\ref{sec:inference_perception}), and the action-perception loop repeats (see also Alg.~\ref{alg:protocol}). Epistemic qualities then emerge from this continual pursuit of evidence \cite{hohwy_conscious_2021}.

In short, we note a distinction in the interpretation of epistemic value between the EFE and the CBFE. In the EFE \eqref{eq:EFE_ep_ex}, epistemic value is directly related with a mutual information term between states and outcomes. In the CBFE, the epistemic drive appears to result from a self-evidencing mechanism.

\subsection{Bethe Free Energy Value Decomposition}
The BFE does not permit an interpretation in terms of intrinsic value. When we substitute \eqref{eq:gm_posterior} in the BFE definition of \eqref{eq:BFE} and combine terms, we obtain
\begin{align}
    \B[q; \hat{\bm{u}}] &= \U{q(\bm{y}, \bm{x})}{p(\bm{y}, \bm{x} | \hat{\bm{u}})} + \U{q(\bm{y})}{\tilde{p}(\bm{y})} - \HB{q(\bm{y}, \bm{x})}\notag\\
    &= \KL{q(\bm{y}, \bm{x})}{p(\bm{y}, \bm{x} | \hat{\bm{u}})} - \E{q(\bm{y})}{\log \tilde{p}(\bm{y})}\notag\\
    &= \underbrace{\E{q(\bm{y})}{\KL{q(\bm{x} | \bm{y})}{p(\bm{x} | \bm{y}, \hat{\bm{u}})}}}_{\text{expected posterior divergence}} + \underbrace{\KL{q(\bm{y})}{p(\bm{y} | \hat{\bm{u}})}}_{\text{predictive divergence}} - \underbrace{\E{q(\bm{y})}{\log \tilde{p}(\bm{y})}}_{\text{expected extrinsic value}}\,. \label{eq:BFE_terms}
\end{align}

The intrinsic value term of \eqref{eq:CBFE_in_ex} has been replaced by a predictive divergence in \eqref{eq:BFE_terms}. This term expresses the difference (divergence) in information between the observations as predicted by the model under policy $\hat{\bm{u}}$, and variational distribution about outcomes $q(\bm{y})$. Under optimization of $q(\bm{x}|\bm{y})$, the posterior divergence will vanish for all $\bm{y}$. Without the additional point-mass constraint of \eqref{eq:q_con_fact}, the predictive divergence then no longer quantifies the information difference between uncertainty (predictive distribution) and certainty (predicted outcomes). As a result, the BFE lacks the self-evidencing qualities and resulting epistemic drive of the CBFE, as we will further illustrate in Sec.~\ref{sec:example_application} and \ref{sec:planning_t_maze}.

\subsection{Example Application}
\label{sec:example_application}
We illustrate our interpretation of \eqref{eq:CBFE_in_ex} and \eqref{eq:BFE_terms} by a minimal example model. We consider a two-armed bandit, where an agent chooses between two levers, $u \in \{0, 1\}$. Each lever offers a distinct probability for observing an outcome $y \in \{0, 1\}$. Specifically, choosing $\hat{u}=0$ will offer a $0.5$ probability for observing $\hat{y}=0$ (ignorant policy), wheres choosing $\hat{u}=1$ will always lead to the observation $\hat{y}=0$ (informative policy). We do not equip the agent with any external preference (there is no extrinsic reward). The agent's factorized model then becomes
\begin{align}
    f(y | u) &= p(y = a_i | u = a_j) = A_{ij}\,, \label{eq:example_model}
\end{align}
with $a = (0, 1)^T$ and the conditional probability matrix
\begin{align}
    A = \begin{pmatrix}
        0.5 & 1\\
        0.5 & 0
    \end{pmatrix}\,.
\end{align}

The BFE then follows as
\begin{align}
    \B[q; \hat{u}] &= \U{q(y)}{p(y | \hat{u})} - \H{q(y)}\,. \label{eq:BFE_example}
\end{align}
The CBFE additionally constrains $q(y) = \delta(y - \hat{y})$, and as a result corresponds directly with the negative intrinsic value term of \eqref{eq:CBFE_in_ex}, as$^{\text{\ref{ftn:entropy}}}$
\begin{align}
    \B(\hat{y}, \hat{u}) &= \U{\delta(y - \hat{y})}{p(y | \hat{u})} - \H{\delta(y - \hat{y})}\notag\\
    &= -\log p(\hat{y} | \hat{u})\,. \label{eq:CBFE_example}
\end{align}

\begin{figure}[ht]
    \hfill
    \begin{minipage}[b]{0.45\textwidth}
        \centering
        \begin{tikzpicture}
            [node distance=20mm,auto]

            \node[box] (T) {$\operatorname{T}$};
            \node[clamped, above of=T, node distance=13mm] (u) {};
            \node[clamped, left of=T, node distance=15mm] (A) {};
            \node[below of=T, node distance=13mm] (optim) {};
            \draw[line] (u) -- node[anchor=west]{$u$} node[anchor=east]{$\downarrow$} (T);
            \draw[line] (A) -- node[anchor=north, pos=0.3]{$A$} node[anchor=south]{$\rightarrow$} (T);    
            \draw[line] (T) -- node[anchor=west, pos=0.6]{$y$} (optim);

            \msg{left}{down}{T}{optim}{0.67}{1}
            \draw[dashed] (-1.9,0.5) rectangle (0.5,-0.5);
            \node[right of=T, node distance=12mm] (p) {$p(y | u)$};
        \end{tikzpicture}
    \end{minipage}
    \begin{minipage}[b]{0.45\textwidth}
        \centering
        \begin{tikzpicture}
            [node distance=20mm,auto]

            \node[box] (T) {$\operatorname{T}$};
            \node[clamped, above of=T, node distance=13mm] (u) {};
            \node[clamped, left of=T, node distance=15mm] (A) {};
            \node[optim, below of=T, node distance=13mm] (optim) {$\delta$};
            \draw[line] (u) -- node[anchor=west]{$u$} node[anchor=east]{$\downarrow$} (T);
            \draw[line] (A) -- node[anchor=north, pos=0.3]{$A$} node[anchor=south]{$\rightarrow$} (T);    
            \draw[line] (T) -- node[anchor=west, pos=0.7]{$y$} (optim);

            \darkmsg{left}{down}{T}{optim}{0.67}{1}
            \draw[dashed] (-1.9,0.5) rectangle (0.5,-0.5);
            \node[right of=T, node distance=12mm] (p) {$p(y | u)$};
        \end{tikzpicture}
    \end{minipage}
    \caption{Message passing schedule for the example model of \eqref{eq:example_model} for the BFE (left) and CBFE (right). The dashed box summarizes the observation model.}
    \label{fig:example}
\end{figure}
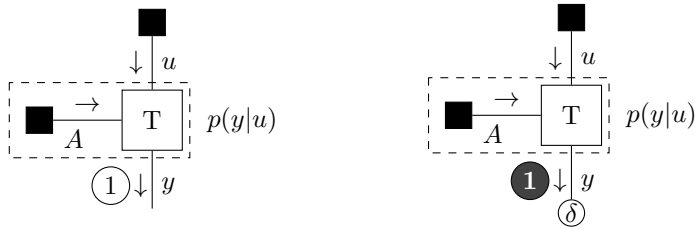

The FFG for the model definition of \eqref{eq:example_model} together with the resulting schedule for optimization of the (C)BFE is drawn in Fig.~\ref{fig:example}.

\begin{table}
\center
\caption{Free energies (in bits) per policy for the example application.}
\label{tbl:example}
\setlength{\extrarowheight}{3pt}
\begin{tabular}{r | l | l }
    Policy                      & BFE & CBFE \\
    \hline
    Ignorant ($\hat{u} = 0$)    & 0   & 1   \\
    Informative ($\hat{u} = 1$) & 0   & 0    \\
\end{tabular}
\end{table}

The results of Table~\ref{tbl:example} show that the BFE does not distinguish between policies. The CBFE however penalizes the ignorant policy ($\hat{u} = 0$), which does not predict precise outcomes. This mechanism thus induces a preference for the informative policy ($\hat{u} = 1$), which does predict precise outcomes. In the following, we will further investigate this behavior in a less trivial setting.

\section{Experimental Setting}
\label{sec:experimental_setting}
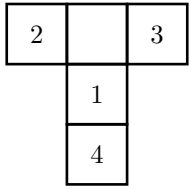
\begin{figure}[ht]
    \hfill
    \begin{center}
        \begin{tikzpicture}
            [node distance=8mm,auto,line width=1pt]

            \node[box] (one) {$1$};
            \node[box, below of=one] (four) {$4$};    
            \node[box, above of=one] (empty) {};
            \node[box, left of=empty] (two) {$2$};
            \node[box, right of=empty] (three) {$3$};
        \end{tikzpicture}
    \end{center}
    \caption{Layout of the T-maze. The agent starts at position 1. The reward is located at either position 2 or 3. Position 4 contains a cue which indicates the reward position.}
    \label{fig:maze_layout}
\end{figure}

A classic experimental setting that investigates epistemic behavior is the T-maze task \cite{friston_active_2015}. The T-maze environment consists of four positions $\mathcal{P} = \{1, 2, 3, 4\}$, as drawn in Fig.~\ref{fig:maze_layout}. The agent starts in position 1, and aims to obtain a reward that resides in either arm 2 or 3, $\mathcal{R}=\{2, 3\}$. The position of the reward is unknown to the agent a priori, and once the agent enters one of the arms it remains there.

In order to learn the position of the reward, the agent first needs to move to position 4, where a cue indicates the reward position. At each position, the agent may observe one of four reward-related outcomes $\mathcal{O} = \{1, 2, 3, 4\}$:
\begin{enumerate}
    \item The reward is indicated to reside at location two (left arm);
    \item The reward is indicated to reside at location three (right arm);
    \item The reward is obtained;
    \item The reward is not obtained.
\end{enumerate}

The key insight is that an epistemic policy would first inspect the cue at position 4 and then move to the indicated arm, whereas a purely goal directed agent would immediately move towards either of the potential goal positions instead of visiting the cue.

\subsection{Generative Model Specification}
We follow \cite{friston_active_2015}, and assume a generative model with discrete states $x_k$,  observations $y_k$ and controls $u_k$. The state $x_k \in \mathcal{P}\times \mathcal{R}$, represents the agent position at time $k$ (four positions, Fig.~\ref{fig:maze_layout}) combined with the reward position (two possibilities). The state vector thus comprises of eight possible realizations. The transition between states is affected by the control $u_k \in \mathcal{P}$, which encodes the agent's attempted next position in the maze. The observation variables $y_k \in \mathcal{O}\times \mathcal{P}$ represent the agent position at time $k$ (four positions) in combination with the reward-related outcome at that position (four possibilities).

The respective state prior, observation model, transition model and goal priors are defined as
\begin{align*}
    p(x_{t-1}) &= \Cat{x_{t-1} | d_{t-1}}\\
    p(y_k=a_i|x_k=b_j) &= A_{ij}\\
    p(x_k=b_i|x_{k-1}=b_j, u_k=\hat{u}_k) &= (B_{\hat{u}_k})_{ij}\\
    \tilde{p}(y_k) &= \Cat{y_k | c_k}\,,
\end{align*}
where $a_i\in \mathcal{O}\times \mathcal{P}$, $b_i \in \mathcal{P}\times \mathcal{R}$, and $b_j \in \mathcal{P}\times \mathcal{R}$.

\begin{figure}[ht]
    \hfill
    \makebox[\textwidth][c]{
        \begin{tikzpicture}
            [node distance=20mm,auto]


            \node[clamped] (D_t_min) {};
            \node[box, right of=D_t_min, node distance=15mm] (cat_D_t_min) {$\mathcal{C}at$};
            \node[box, right of=cat_D_t_min, node distance=27mm] (B_t) {$\operatorname{T}$};
            \node[smallbox, right of=B_t] (eq_t) {$=$};
            \node[box, above of=B_t, node distance=15mm] (mux_t) {$\operatorname{MUX}$};
            \node[clamped, above of=mux_t, node distance=15mm] (u_t) {};
            \node[clamped, left of=mux_t, node distance=15mm] (B_mux_t) {};
            \node[box, below of=eq_t, node distance=15mm] (A_t) {$\operatorname{T}$};
            \node[clamped, left of=A_t, node distance=15mm] (del_A_t) {};
            \node[box, below of=A_t, node distance=20mm] (cat_C_t) {$\mathcal{C}at$};
            \node[clamped, below of=cat_C_t, node distance=15mm] (C_t) {};

            \draw[line] (D_t_min) -- node[anchor=south]{$d_{t-1}$} (cat_D_t_min);
            \draw[line] (cat_D_t_min) -- node[anchor=south]{$x_{t-1}$} (B_t);
            \draw[line] (u_t) -- node[anchor=west]{$u_t$} (mux_t);
            \draw[line] (B_mux_t) -- node[anchor=south]{$\bm{B}$} (mux_t);
            \draw[line] (mux_t) -- node[anchor=west]{$B_{u_t}$} (B_t);
            \draw[line] (B_t) -- (eq_t);
            \draw[line] (eq_t) -- node[anchor=east, pos=0.4]{$x_t$} (A_t);
            \draw[line] (del_A_t) -- node[anchor=north]{$A$} (A_t);
            \draw[line] (A_t) -- node[anchor=west]{$y_t$} (cat_C_t);
            \draw[line] (cat_C_t) -- node[anchor=west]{$c_t$} (C_t);

            \node[box, right of=eq_t, node distance=24mm] (B_t_plus) {$\operatorname{T}$};
            \node[box, above of=B_t_plus, node distance=15mm] (mux_t_plus) {$\operatorname{MUX}$};
            \node[clamped, above of=mux_t_plus, node distance=15mm] (u_t_plus) {};
            \node[clamped, left of=mux_t_plus, node distance=15mm] (B_mux_t_plus) {};
            \coordinate[right of=B_t_plus] (eq_t_plus);
            \node[box, below of=eq_t_plus, node distance=15mm] (A_t_plus) {$\operatorname{T}$};
            \node[clamped, left of=A_t_plus, node distance=15mm] (del_A_t_plus) {};
            \node[box, below of=A_t_plus, node distance=20mm] (cat_C_t_plus) {$\mathcal{C}at$};
            \node[clamped, below of=cat_C_t_plus, node distance=15mm] (C_t_plus) {};

            \draw[line] (eq_t) -- (B_t_plus);
            \draw[line] (u_t_plus) -- node[anchor=west]{$u_{t+1}$} (mux_t_plus);
            \draw[line] (B_mux_t_plus) -- node[anchor=south]{$\bm{B}$} (mux_t_plus);
            \draw[line] (mux_t_plus) -- node[anchor=west]{$B_{u_{t+1}}$} (B_t_plus);
            \draw[line] (B_t_plus) -- (eq_t_plus);
            \draw[line] (eq_t_plus) -- node[anchor=west]{$x_{t+1}$} (A_t_plus);
            \draw[line] (del_A_t_plus) -- node[anchor=north]{$A$} (A_t_plus);
            \draw[line] (A_t_plus) -- node[anchor=west]{$y_{t+1}$} (cat_C_t_plus);
            \draw[line] (cat_C_t_plus) -- node[anchor=west]{$c_{t+1}$} (C_t_plus);

            \draw[dashed] (-0.4,0.6) rectangle (2.1,-0.6);
            \draw[dashed] (2.3,2.1) rectangle (5.1,-0.6);
            \draw[dashed] (6.7,2.1) rectangle (9.8,-0.6);
            \draw[dashed] (4.3,-0.9) rectangle (6.8,-2.1);
            \draw[dashed] (8.7,-0.9) rectangle (11.2,-2.1);
            \draw[dashed] (5.6,-2.9) rectangle (6.8,-5.4);
            \draw[dashed] (10.0,-2.9) rectangle (11.5,-5.4);
            \node (prior) at (0.5,-1.0) {$p(x_{t-1})$};
            \node (trans_t) at (2.7,2.5) {$p(x_t | x_{t-1}, u_t)$};
            \node (trans_t_plus) at (6.9,2.5) {$p(x_{t+1} | x_t, u_{t+1})$};
            \node (obs_t) at (4.0,-2.5) {$p(y_t | x_t)$};
            \node (obs_t_plus) at (9.0,-2.5) {$p(y_{t+1} | x_{t+1})$};
            \node (goal_t) at (5.0,-4.0) {$\tilde{p}(y_t)$};
            \node (goal_t_plus) at (9.2,-4.0) {$\tilde{p}(y_{t+1})$};
        \end{tikzpicture}
    }
    \caption{Forney-style factor graph of the generative model for the T-maze. The $\operatorname{MUX}$ nodes select the transition matrix as determined by the control variable. Dashed boxes summarize the indicated distributions.}
    \label{fig:maze_model}
\end{figure}
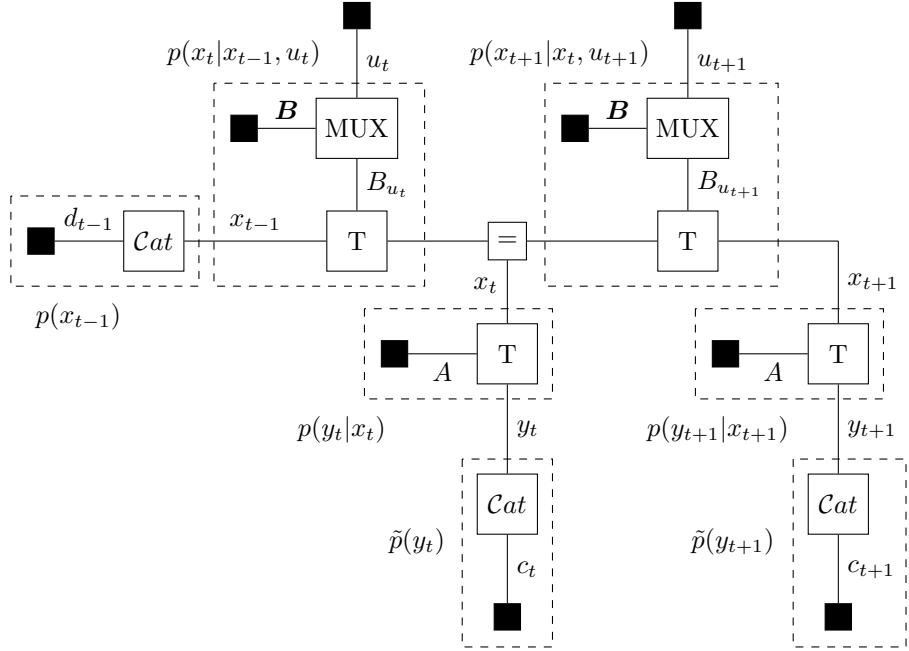

The agent plans two steps ahead ($T=2$), for which the FFG is drawn in Fig.~\ref{fig:maze_model}.

\subsection{Parameter Assignments}
We start the simulation at $t=1$, and assume that the initial position is known, namely we start at position 1 (Fig.~\ref{fig:maze_layout}). However, the reward position is unknown a priori. This prior information is encoded by the initial state probability vector
\begin{align*}
    d_0 &= (1, 0, 0, 0)^{\T} \otimes (0.5, 0.5)^{\T}\,,
\end{align*}
where $\otimes$ denotes the Kronecker product.

The transition matrix $B_{u_k}$ encodes the state transitions (from column-index to row-index), as
\begin{align*}
    B_1 &=
    \begin{pmatrix}
        1 & 0 & 0 & 1\\
        0 & 1 & 0 & 0\\
        0 & 0 & 1 & 0\\
        0 & 0 & 0 & 0
    \end{pmatrix} \otimes I_2,\,
    B_2 =
    \begin{pmatrix}
        0 & 0 & 0 & 0\\
        1 & 1 & 0 & 1\\
        0 & 0 & 1 & 0\\
        0 & 0 & 0 & 0
    \end{pmatrix} \otimes I_2,\,\\
    B_3 &=
    \begin{pmatrix}
        0 & 0 & 0 & 0\\
        0 & 1 & 0 & 0\\
        1 & 0 & 1 & 1\\
        0 & 0 & 0 & 0
    \end{pmatrix} \otimes I_2,\,
    B_4 =
    \begin{pmatrix}
        0 & 0 & 0 & 0\\
        0 & 1 & 0 & 0\\
        0 & 0 & 1 & 0\\
        1 & 0 & 0 & 1
    \end{pmatrix} \otimes I_2\,.
\end{align*}
The control affects the agent position, but not the reward position. Therefore, Kronecker products with the two-dimensional unit matrix $I_2$ ensure that the transitions are duplicated for both possible reward positions. Note that positions $2$ and $3$ (the reward arms) are attracting states, since none of the transition matrices allow a transition away from these positions. This means that although it is possible to propose any control at any time, not all controls will move the agent to its attempted position. We denote the collection of transition matrices by $\bm{B} = \{B_1, B_2, B_3, B_4\}$.

The observed outcome depends on the position of the agent. The position-dependent observation matrices specify how observations follow, given the current position of the agent (subscripts) and the reward position (columns), as
\begin{align}
    A_1 = 
    \begin{pmatrix}
        0.5 & 0.5\\
        0.5 & 0.5\\
        0 & 0\\
        0 & 0
    \end{pmatrix},\,
    A_2 = 
    \begin{pmatrix}
        0 & 0\\
        0 & 0\\
        \alpha & 1-\alpha\\
        1-\alpha & \alpha
    \end{pmatrix},\,
    A_3 = 
    \begin{pmatrix}
        0 & 0\\
        0 & 0\\
        1-\alpha & \alpha\\
        \alpha & 1-\alpha
    \end{pmatrix},\,
    A_4 = 
    \begin{pmatrix}
        1 & 0\\
        0 & 1\\
        0 & 0\\
        0 & 0
    \end{pmatrix},\, \label{eq:A_alpha}
\end{align}
with reward probability $\alpha$. The columns of these position-dependent observation matrices represent the two possibilities for the reward position. The position-dependent observation matrices combine into the complete block-diagonal, 16-by-8 observation matrix
\begin{align*}
    A = A_1 \oplus A_2 \oplus A_3 \oplus A_4\,,
\end{align*}
where $\oplus$ denotes the direct sum (i.e. block-diagonal concatenation).

The goal prior depends upon the future time,
\begin{subequations}
\label{eq:C}
\begin{align}
    c_1 &= (0.25, 0.25, 0.25, 0.25)^{\T} \otimes (0.25, 0.25, 0.25, 0.25)^{\T}\\
    c_k &= \sigma\!\left((0, 0, c, -c)^{\T} \otimes (1, 1, 1, 1)^{\T}\right) \text{ for $k > 1$}\,,
\end{align}
\end{subequations}
with reward utility $c$, and $\sigma$ the soft-max function where $\sigma(s)_i = \frac{\operatorname{exp}(s_i)}{\sum_j \operatorname{exp}(s_j)}$. The flat prior $c_1$ encodes a lack of external preference at $t=1$, while $c_k$ for $k>1$ encodes a preference for observing rewards at subsequent times. This effectively removes the goal prior for the first move ($t=1$), while in subsequent moves the agent is rewarded for extrinsically rewarding states \cite{van_de_laar_simulating_2018}.

\section{Inference for Planning}
\label{sec:planning_t_maze}
In this simulation we compare the behavior of a CBFE agent to the behavior of a reference BFE agent (without point-mass constraints). We consider given policies $\hat{\bm{u}}$, and the optimal CBFE as a function of those policies
\begin{align}
    \B_{\hat{\bm{y}}}^*(\hat{\bm{u}}) = \min_{q, \hat{\bm{y}}} \B[q; \hat{\bm{y}}, \hat{\bm{u}}]\,, \label{eq:F_maze_con}
\end{align}
where the $\hat{\bm{y}}$ subscript indicates the explicit inclusion of point-mass constraints.

The unconstrained BFE represents the objective where the future observation variables $\bm{y}$ are not point-mass constrained by their potential outcomes $\hat{\bm{y}}$. The unconstrained agent will therefore optimize the joint belief over state and future observation variables rather than potential outcomes, as
\begin{align}
    \B^*(\hat{\bm{u}}) &= \min_{q} \B[q; \hat{\bm{u}}]\,. \label{eq:F_maze_ref}
\end{align}

We will evaluate the BFE, CBFE and EFE for all sixteen ($T=2$) possible candidate policies $\hat{\bm{u}} \in \mathcal{U}\times \mathcal{U}$. We consider several scenarios with varying reward probabilities $\alpha$ \eqref{eq:A_alpha} and reward utilities $c$ \eqref{eq:C}.

In the current section we do not (yet) consider the interaction of the agent with the environment. In other words, actions from optimal policy $\hat{\bm{u}}^*$ are not (yet) executed; we are purely interested in the inference for planning itself, and the resulting free energy values as a function of the candidate policies \eqref{eq:F_maze_con}.

\subsection{Message Passing Schedule for Planning}
\label{sec:message_passing_schedule}
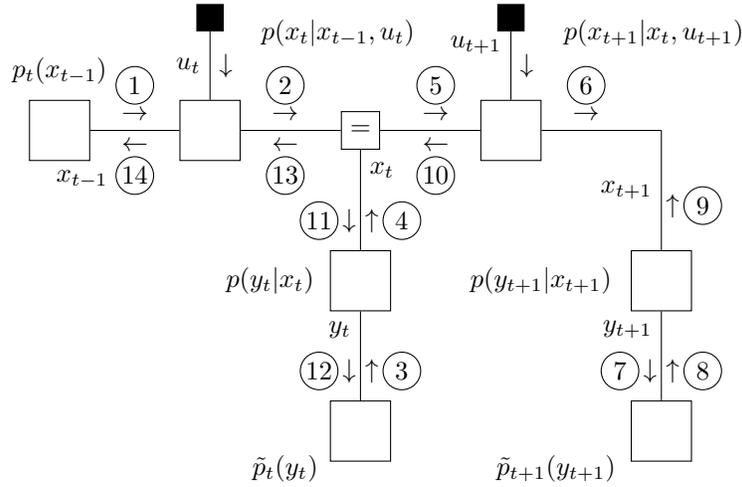
\begin{figure}[ht]
    \hfill
    \begin{center}
        \begin{tikzpicture}
            [node distance=20mm,auto]

            \node[box] (D_t_min) {};
            \node[box, right of=D_t_min] (B_t) {};
            \node[smallbox, right of=B_t] (eq_t) {$=$};
            \node[clamped, above of=B_t, node distance=15mm] (u_t) {};
            \node[box, below of=eq_t, node distance=20mm] (A_t) {};
            \node[box, below of=A_t, node distance=20mm] (C_t) {};

            \draw[line] (D_t_min) -- node[anchor=north, xshift=-7mm, yshift=-4mm]{$x_{t-1}$} (B_t);
            \draw[line] (u_t) -- node[anchor=east]{$u_t$} node[anchor=west]{$\downarrow$} (B_t);
            \draw[line] (B_t) -- (eq_t);
            \draw[line] (eq_t) -- node[anchor=west, pos=0.2]{$x_t$} (A_t);
            \draw[line] (A_t) -- node[anchor=east, pos=0.2]{$y_t$} (C_t);

            \node[box, right of=eq_t] (B_t_plus) {};
            \coordinate[right of=B_t_plus] (eq_t_plus);
            \node[clamped, above of=B_t_plus, node distance=15mm] (u_t_plus) {};
            \node[box, below of=eq_t_plus, node distance=20mm] (A_t_plus) {};
            \node[box, below of=A_t_plus, node distance=20mm] (C_t_plus) {};

            \draw[line] (eq_t) -- (B_t_plus);
            \draw[line] (u_t_plus) -- node[anchor=east, pos=0.2]{$u_{t+1}$} node[anchor=west]{$\downarrow$} (B_t_plus);
            \draw[line] (B_t_plus) -- (eq_t_plus);
            \draw[line] (eq_t_plus) -- node[anchor=east]{$x_{t+1}$} (A_t_plus);
            \draw[line] (A_t_plus) -- node[anchor=east, pos=0.2]{$y_{t+1}$} (C_t_plus);

            \msg{up}{right}{D_t_min}{B_t}{0.5}{1}
            \msg{up}{right}{B_t}{eq_t}{0.5}{2}
            \msg{right}{up}{A_t}{eq_t}{0.4}{4}
            \msg{up}{right}{eq_t}{B_t_plus}{0.5}{5}
            \msg{up}{right}{B_t_plus}{eq_t_plus}{0.5}{6}
            \msg{right}{up}{A_t_plus}{eq_t_plus}{0.5}{9}
            \msg{down}{left}{eq_t}{B_t_plus}{0.5}{10}
            \msg{left}{down}{A_t}{eq_t}{0.4}{11}
            \msg{down}{left}{B_t}{eq_t}{0.5}{13}
            \msg{down}{left}{D_t_min}{B_t}{0.5}{14}

            \msg{left}{down}{A_t}{C_t}{0.6}{12}
            \msg{right}{up}{A_t}{C_t}{0.6}{3}
            \msg{left}{down}{A_t_plus}{C_t_plus}{0.6}{7}
            \msg{right}{up}{A_t_plus}{C_t_plus}{0.6}{8}

            \node (prior) at (0.0,0.8) {$p_t(x_{t-1})$};
            \node (trans_t) at (3.7,1.3) {$p(x_t | x_{t-1}, u_t)$};
            \node (trans_t_plus) at (7.9,1.3) {$p(x_{t+1} | x_t, u_{t+1})$};
            \node (obs_t) at (2.8,-2.0) {$p(y_t | x_t)$};
            \node (obs_t_plus) at (6.4,-2.0) {$p(y_{t+1} | x_{t+1})$};
            \node (goal_t) at (3.0,-4.5) {$\tilde{p}_t(y_t)$};
            \node (goal_t_plus) at (6.6,-4.5) {$\tilde{p}_{t+1}(y_{t+1})$};
        \end{tikzpicture}
    \end{center}
    \caption{Message passing schedule for planning in the T-maze with the BFE.}
    \label{fig:maze_schedule_BFE}
\end{figure}

\begin{figure}[ht]
    \hfill
    \begin{center}
        \begin{tikzpicture}
            [node distance=20mm,auto]

            \node[box] (D_t_min) {};
            \node[box, right of=D_t_min] (B_t) {};
            \node[smallbox, right of=B_t] (eq_t) {$=$};
            \node[clamped, above of=B_t, node distance=15mm] (u_t) {};
            \node[box, below of=eq_t, node distance=20mm] (A_t) {};
            \node[box, below of=A_t, node distance=25mm] (C_t) {};

            \draw[line] (D_t_min) -- node[anchor=north, xshift=-7mm, yshift=-4mm]{$x_{t-1}$} (B_t);
            \draw[line] (u_t) -- node[anchor=east]{$u_t$} node[anchor=west]{$\downarrow$} (B_t);
            \draw[line] (B_t) -- (eq_t);
            \draw[line] (eq_t) -- node[anchor=west, pos=0.2]{$x_t$} (A_t);
            \draw[line] (A_t) -- node[anchor=east, pos=0.8]{$y_t$} (C_t);

            \node[box, right of=eq_t] (B_t_plus) {};
            \coordinate[right of=B_t_plus] (eq_t_plus);
            \node[clamped, above of=B_t_plus, node distance=15mm] (u_t_plus) {};
            \node[box, below of=eq_t_plus, node distance=20mm] (A_t_plus) {};
            \node[box, below of=A_t_plus, node distance=25mm] (C_t_plus) {};

            \draw[line] (eq_t) -- (B_t_plus);
            \draw[line] (u_t_plus) -- node[anchor=east, pos=0.2]{$u_{t+1}$} node[anchor=west]{$\downarrow$} (B_t_plus);
            \draw[line] (B_t_plus) -- (eq_t_plus);
            \draw[line] (eq_t_plus) -- node[anchor=east]{$x_{t+1}$} (A_t_plus);
            \draw[line] (A_t_plus) -- node[anchor=east, pos=0.8]{$y_{t+1}$} (C_t_plus);

            \node[optim, below of=A_t, node distance=12mm] (optim_t) {$\delta$};
            \node[optim, below of=A_t_plus, node distance=12mm] (optim_t_plus) {$\delta$};

            \msg{up}{right}{D_t_min}{B_t}{0.5}{1}
            \msg{up}{right}{B_t}{eq_t}{0.5}{2}
            \darkmsg{right}{up}{A_t}{eq_t}{0.4}{3}
            \msg{up}{right}{eq_t}{B_t_plus}{0.5}{4}
            \msg{up}{right}{B_t_plus}{eq_t_plus}{0.5}{5}
            \darkmsg{right}{up}{A_t_plus}{eq_t_plus}{0.5}{6}
            \msg{down}{left}{eq_t}{B_t_plus}{0.5}{7}
            \msg{left}{down}{A_t}{eq_t}{0.4}{8}
            \msg{down}{left}{B_t}{eq_t}{0.5}{9}
            \msg{down}{left}{D_t_min}{B_t}{0.5}{10}

            \darkmsg{left}{down}{A_t}{C_t}{0.3}{11}
            \msg{right}{up}{A_t}{C_t}{0.7}{12}
            \darkmsg{left}{down}{A_t_plus}{C_t_plus}{0.3}{13}
            \msg{right}{up}{A_t_plus}{C_t_plus}{0.7}{14}

            \node (prior) at (0.0,0.8) {$p_t(x_{t-1})$};
            \node (trans_t) at (3.7,1.3) {$p(x_t | x_{t-1}, u_t)$};
            \node (trans_t_plus) at (7.9,1.3) {$p(x_{t+1} | x_t, u_{t+1})$};
            \node (obs_t) at (2.8,-2.0) {$p(y_t | x_t)$};
            \node (obs_t_plus) at (6.4,-2.0) {$p(y_{t+1} | x_{t+1})$};
            \node (goal_t) at (3.0,-4.5) {$\tilde{p}_t(y_t)$};
            \node (goal_t_plus) at (6.6,-4.5) {$\tilde{p}_{t+1}(y_{t+1})$};
        \end{tikzpicture}
    \end{center}
    \caption{Message passing schedule for planning in the T-maze with the CBFE.}
    \label{fig:maze_schedule_CBFE}
\end{figure}
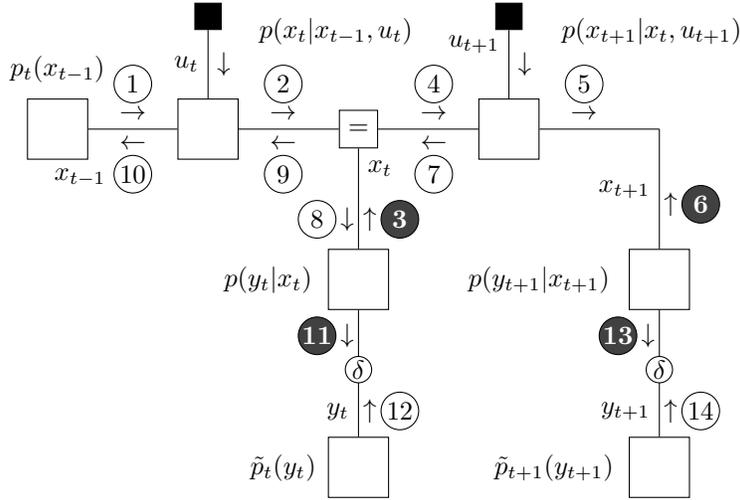

The message passing schedules for planning are drawn in Fig.~\ref{fig:maze_schedule_BFE} (BFE) and \ref{fig:maze_schedule_CBFE} (CBFE), where light messages are computed by sum-product (SP) message passing updates \cite{loeliger_introduction_2004}, and dark messages by variational message passing updates \cite{dauwels_variational_2007}. An overview of message passing updates for discrete nodes can be found in \cite[App.~A]{van_de_laar_automated_2019}.


For the CBFE, the posterior beliefs associated with the observation variables are constrained by point-mass (Dirac-delta) distributions, see \eqref{eq:q_con_fact}, and the corresponding potential outcomes are optimized for. The message passing optimization scheme is derived from first principles in \cite{senoz_variational_2021}. In order to obtain a new value, e.g. $\hat{y}_t$, messages $\smalldarkcircled{\!11}$ and $\smallcircled{\!12}$ are multiplied. The mode of the product then becomes the new value $\hat{y}_t$, which is used to construct the belief $q(y_t) = \delta(y_t - \hat{y}_t)$. The updated belief is subsequently used in the next iteration to compute $\smalldarkcircled{3}$. The resulting iterative expectation maximization (EM) procedure initializes values for all $\hat{y}_k$, and is performed using message passing according to \cite{dauwels_expectation_2005}. Interestingly, where optimization of the EFE is performed by a forward-only procedure (see Appendix~A), optimization of the (C)BFE, as illustrated in Fig.~\ref{fig:maze_schedule_BFE} and \ref{fig:maze_schedule_CBFE}, also includes a complete backward (smoothing) pass over the model.

\subsection{Inference Results for Planning}
\label{sec:inference_results_planning}
\begin{figure}[ht]
    \centering
    \makebox[\textwidth][c]{\includegraphics{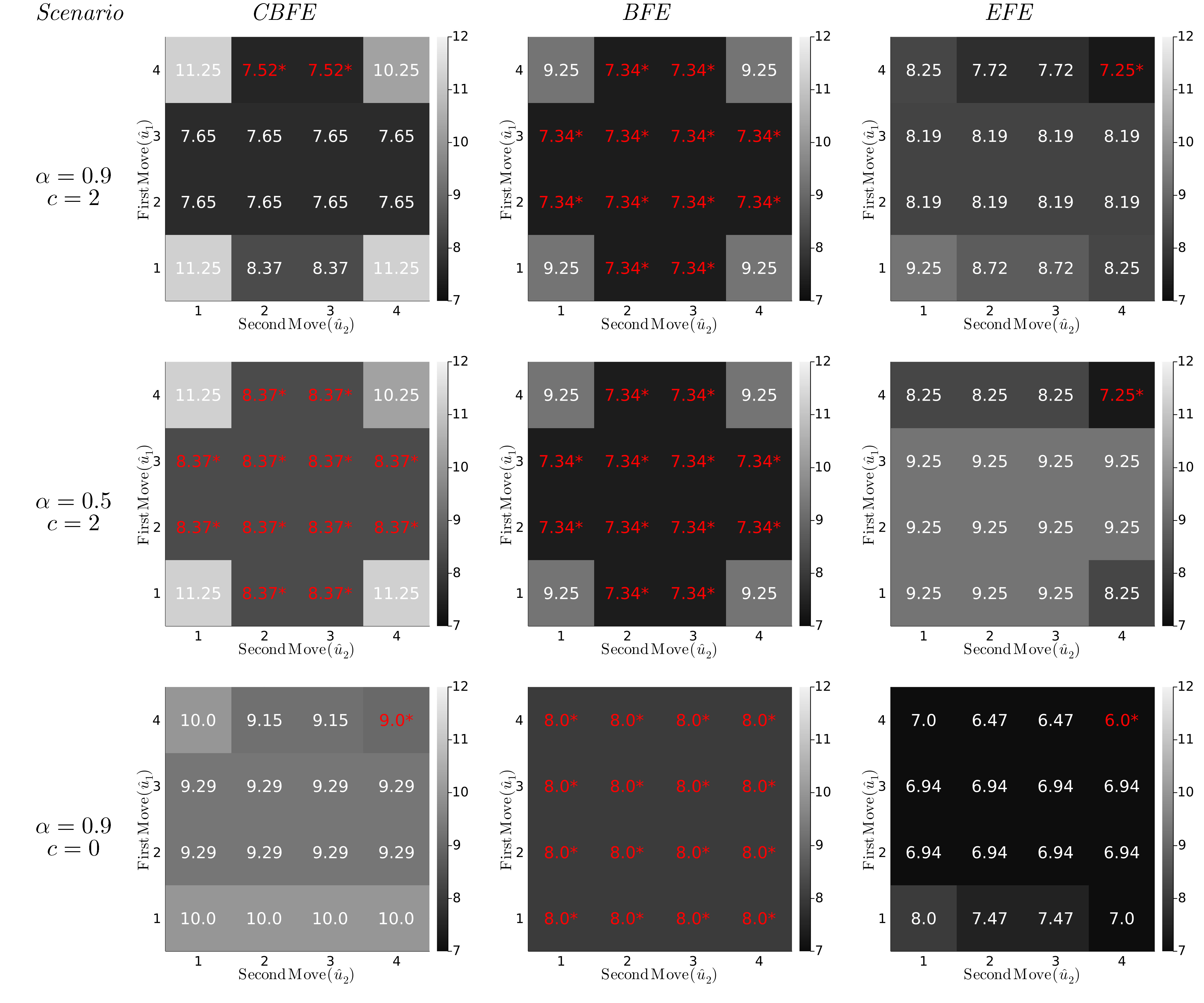}}
    \caption{(Constrained) Bethe Free Energies ((C)BFE) and Expected Free Energies (EFE) (in bits) for the T-maze policies under varying parameter settings. Each diagram plots the minimized free energy values for all possible policies (lookahead $T=2$), with the first move on the vertical axis and the second move on the horizontal axis. For example, the cell in row 4, column 3 represents the policy $\hat{\bm{u}}=(4,3)$, which first moves to position 4 and then to position 3. The values for the optimal policies $\hat{\bm{u}}^*$ are annotated red with an asterisk.}
    \label{fig:maze_results}
\end{figure}

Optimization of the (C)BFE by message passing is performed with \textrm{ForneyLab}\footnote{\textrm{ForneyLab} is available at \url{https://github.com/biaslab/ForneyLab.jl}.}\footnote{Simulation source code is available at \url{https://biaslab.github.io/materials/epistemic_search.zip}.} version 0.11.4 \cite{cox_factor_2019}. Free energies for planning, for three different agents and T-maze scenarios, are plotted in Fig.~\ref{fig:maze_results}. The distinct agents optimize the CBFE, BFE and EFE, respectively. We summarize the most important observations below.

The first column of diagrams in Fig.~\ref{fig:maze_results} shows the results for the CBFE agent, for varying scenarios.
\begin{itemize}
\item The first scenario for the CBFE agent (upper left diagram) imposes a likely reward ($\alpha=0.9$) and positive reward utility ($c=2$). In this scenario, the CBFE agent prefers the informative policies (4,2) and (4,3), where the agent seeks the cue in the first move and the reward in the second move. An epistemic (information seeking) agent would prefer these policies in this scenario.
\item In the upper left diagram, note the lack of preference between position 2 and 3 in the second move. Because the policy is not yet executed (moves are only planned), the true reward location remains unknown. Therefore, both of these informative policies are on equal footing.
\end{itemize}

The second column of diagrams shows the results for the BFE agent.
\begin{itemize}
\item In every scenario, the BFE agent fails to distinguish between the majority of ignorant (first move to 1), informative (first move to 4) and greedy policies (first move to 2 or 3). These policy preferences do not correspond with the anticipated preferences of an epistemic agent.
\item Comparing the BFE with the CBFE results, we observe that the point-mass constraint on potential outcomes induces a differentiation between ignorant, informative and greedy policies.
\item More specifically, the third scenario (third row of diagrams) removes the extrinsic value of reward ($c=0$). While the CBFE still differentiates between ignorant, informative and greedy policies, the BFE agent exhibits a total lack of preference.
\item The second scenario (second row of diagrams) removes the value of information about the reward position ($\alpha=0.5$). This scenario thus renders the cue worthless. The BFE agent appears insusceptible to a change in the epistemic $\alpha$ parameter.
\end{itemize}
Taken together, these observations support the interpretation of the BFE as a purely extrinsically driven objective (Sec.~\ref{sec:epistemic_contribution}).

The third column of diagrams produces the results for an EFE agent, as implemented in accordance with \cite{friston_active_2015}, see also Appendix~A.
\begin{itemize}
\item In all scenarios, the EFE agent exhibits a consistent preference for the (4,4) policy. Compared to the CBFE agent, the EFE agent fails to plan ahead to obtain future reward after observing the cue.
\item As we will see in Sec.~\ref{sec:interactive_simulation}, the EFE agent only infers a preference for a reward arm (position 2 or 3) after \emph{execution} of the first move to the cue position. In contrast, the CBFE agent predicts the impact of information and plans accordingly.
\item The second scenario (middle row) provides an informative cue, but removes the possibility to exploit that information ($\alpha=0.5$). Interestingly, the EFE agent still moves to the que position (for the sake of getting information), whereas the CBFE agent expresses ambivalence under an inoperable cue.
\end{itemize}

\FloatBarrier
\subsection{Results for CBFE Value Decomposition}
\begin{figure}[ht]
    \centering
    \makebox[\textwidth][c]{\includegraphics{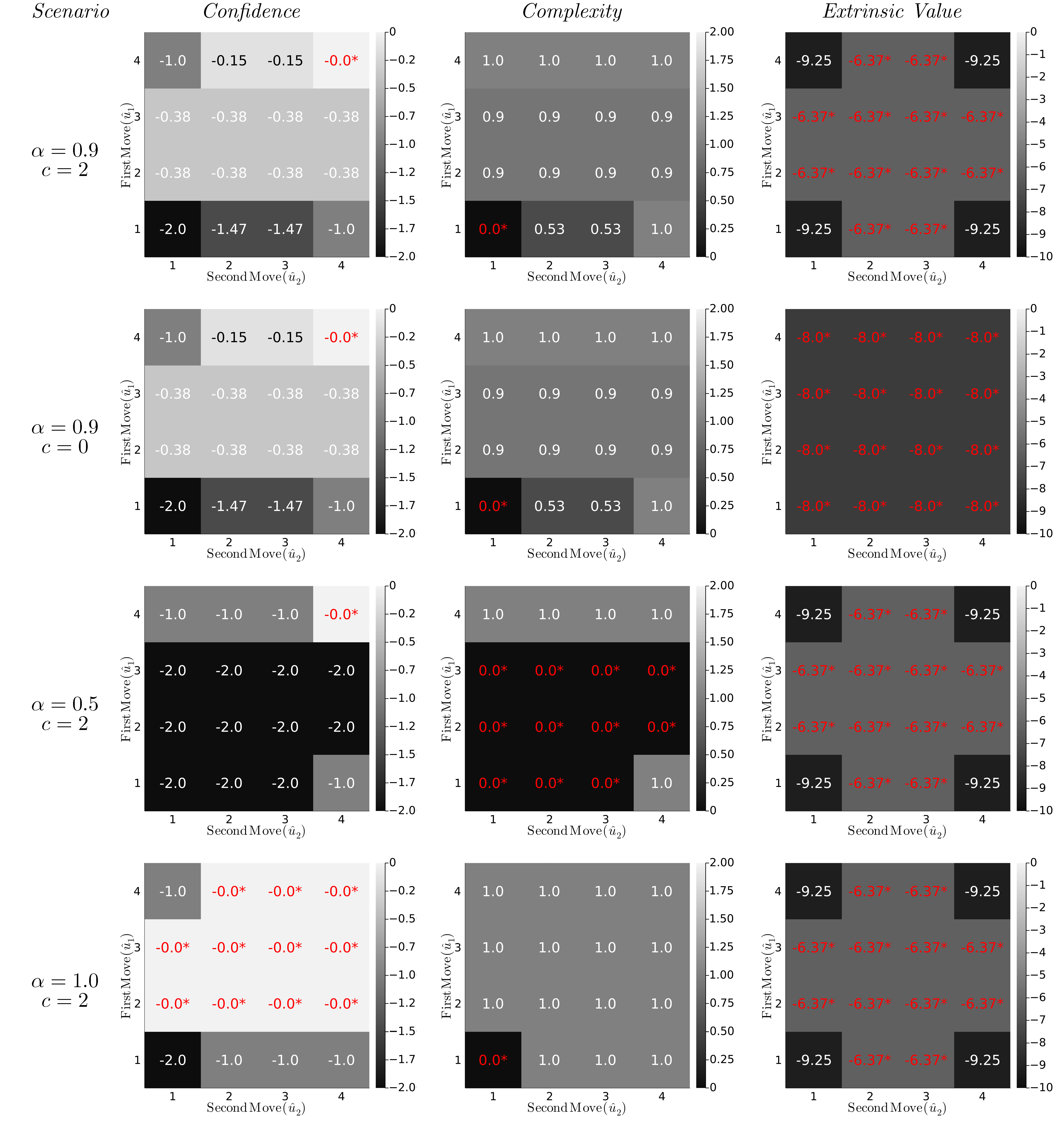}}
    \caption{Confidence, complexity and extrinsic value contributions (in bits) to the Constrained Bethe Free Energy \eqref{eq:CBFE_confidence_complexity} for the T-maze policies (lookahead $T=2$) under varying parameter settings. Optimal values are indicated red with an asterisk.}
    \label{fig:maze_values}
\end{figure}

Simulated values for the CBFE decomposition \eqref{eq:CBFE_confidence_complexity} in the T-maze application are shown in Fig.~\ref{fig:maze_values}, for four different T-maze scenarios. We summarize the most important observations below.

The first column of diagrams in Fig~\ref{fig:maze_values} represents the confidence \eqref{eq:CBFE_in_ex} of the CBFE objective for all (planned) policies. The confidence prefers (or ties) the most informative policy (4,4) for all scenarios.
\begin{itemize}
\item In the first three scenarios (first three rows of diagrams), all policies other than (4,4) dismiss the opportunity to obtain full information about outcomes on two occasions ($T=2$). This is reflected by a negative confidence value, which measures the average rejected information in bits. For example, the policy (1,1) rejects two possibilities to obtain $1$ bit of information, leading to a confidence of $-2$.
\item A change in the external value parameter $c$ does not affect the confidence, which supports the interpretation of the confidence as an intrinsic quantity \eqref{eq:epistemic_balance}.
\item In the final scenario, the greedy policies (moving first to position 2 or 3) are on equal footing with the informative policies (moving first to 4). This is because in the final scenario, visiting position 2 or 3 offers the same amount of information (namely, complete certainty) about the reward position, as would visiting the cue position.
\end{itemize}

The complexity (second column of diagrams) opposes changes in state beliefs that are unwarranted by the policy-induced state transitions, and guards against premature convergence of the state precision (Table~\ref{tbl:CBFE}). As a result, the complexity prefers (or ties) the most conservative policy (1,1) for all scenarios.
\begin{itemize}
\item The complexity is unaffected by changes in utility (similar to the confidence), which supports the interpretation of the complexity as an intrinsic quantity \eqref{eq:epistemic_balance}.
\item In the third scenario ($\alpha=0.5$), the greedy policies become tied in complexity with (most of) the ignorant policies. Because neither visiting a reward arm nor remaining at the initial position offers any useful information about the reward position, the state belief remains unaltered, and these policies incur no complexity penalty.
\end{itemize}

The extrinsic value (third column of diagrams) represents the value of external reward, and leads the agent to pursue extrinsically rewarding states.
\begin{itemize}
\item The extrinsic value is unaffected by changes in the epistemic reward probability parameter $\alpha$, which supports the interpretation of the extrinsic value as an externally determined quantity.
\item In the second scenario the reward utility vanishes ($c=0$), and the extrinsic value becomes indifferent about policies.
\end{itemize}

\FloatBarrier
\section{Interactive Simulation}
\label{sec:interactive_simulation}
In this section we compare the resulting behavior of the CBFE agent with a traditional EFE agent, in \emph{interaction} with a simulated environment.

\subsection{Experimental Protocol}
\label{sec:experimental_protocol}
The experimental protocol governs how the agent interacts with its environment. In our protocol, the action and outcome at time $t$ are the only quantities that are exchanged between the agent and the environment (generative process). The task of the agent is then to plan for actions that lead the agent to desired states. We adapt the experimental protocol of \cite{van_de_laar_simulating_2019} for the purpose of the current simulation. We write the model $f_t$ with a time-subscript to indicate the time-dependent statistics of the state prior as a result of the perceptual process (Sec.~\ref{sec:inference_perception}). The experimental protocol (Alg.~\ref{alg:protocol}) then consists of five steps per time $t$.

\begin{algorithm}
\caption{Experimental protocol.}
\label{alg:protocol}
\begin{algorithmic}
\STATE {Given a model $f_1$ with initial state and goal priors}
\FOR{$t=1$ to $N$}
\STATE {\phantom{$f_{t+1}=$}\llap{$\hat{\bm{u}}^*_t=$} \rlap{\textbf{plan}$(f_t)$}\phantom{\textbf{slide}$(\hat{u}^*_t, \hat{y}_t)$} \quad\# \textit{Execute the planning algorithm}}
\STATE {\phantom{$f_{t+1}=$}\llap{$\hat{u}^*_t=$} \rlap{\textbf{act}$(\hat{\bm{u}}^*_t)$}\phantom{\textbf{slide}$(\hat{u}^*_t, \hat{y}_t)$} \quad\# \textit{Select the first action}}
\STATE {\phantom{$f_{t+1}=$}~\rlap{\textbf{execute}$(\hat{u}^*_t)$}\phantom{\textbf{slide}$(\hat{u}^*_t, \hat{y}_t)$} \quad\# \textit{Execute the action in the simulated environment}}
\STATE {\phantom{$f_{t+1}=$}\llap{$\hat{y}_t=$} \rlap{\textbf{observe}$()$}\phantom{\textbf{slide}$(\hat{u}^*_t, \hat{y}_t)$} \quad\# \textit{Observe the new environmental outcome}}
\STATE {$f_{t+1}=$ \textbf{slide}$(\hat{u}^*_t, \hat{y}_t)$ \quad\# \textit{Prepare the model for the next iteration}}
\ENDFOR
\end{algorithmic}
\end{algorithm}

The \textbf{plan} step solves the inference for planning (Sec.~\ref{sec:inference_planning}), and returns the active policy $\hat{\bm{u}}^*_t$ that represents the (believed) optimal sequence of future controls. In the \textbf{act} step, the first action $\hat{u}^*_t$ is picked from the policy. The \textbf{execute} step then subsequently executes this action in the simulated environment. Execution will alter the state of the environment. In the \textbf{observe} step, the environment responds with a new observation $\hat{y}_t$. Given the action and resulting observation, the \textbf{slide} step then solves the inference for perception (Sec.~\ref{sec:inference_perception}) and prepares the model for the next step.

Inference for the \textbf{slide} step is illustrated in Fig.~\ref{fig:slide_schedule}, where message $\smallcircled{3}$ propagates an observed outcome $\hat{y}_t$, and where message $\smallcircled{5}$ summarizes the information contained within in the dashed box. Only the dashed sub-model is relevant to the \textbf{slide} step, that is, beliefs about the future do not influence $\smallcircled{5}$. After computation, message $\smallcircled{5}$ is normalized, and the resulting state posterior $q^{*}(x_t)$ is subsequently used as a prior to construct the model $f_{t+1}$ for the next time-step, see also \cite{van_de_laar_simulating_2019}.
\begin{figure}[ht]
    \hfill
    \makebox[\textwidth][c]{
        \begin{tikzpicture}
            [node distance=20mm,auto]

            \node[box] (D_t_min) {};
            \node[box, right of=D_t_min] (B_t) {};
            \node[smallbox, right of=B_t] (eq_t) {$=$};
            \node[clamped, above of=B_t, node distance=15mm] (u_t) {};
            \node[box, below of=eq_t, node distance=20mm] (A_t) {};
            \node[box, below of=A_t, node distance=25mm] (C_t) {};

            \draw[line] (D_t_min) -- node[anchor=north, xshift=-7mm, yshift=-4mm]{$x_{t-1}$} (B_t);
            \draw[line] (u_t) -- node[anchor=east]{$u_t$} node[anchor=west]{$\downarrow$} (B_t);
            \draw[line] (B_t) -- (eq_t);
            \draw[line] (eq_t) -- node[anchor=west, pos=0.2]{$x_t$} (A_t);
            \draw[line] (A_t) -- node[anchor=east, pos=0.2]{$y_t$} (C_t);

            \node[box, right of=eq_t] (B_t_plus) {};
            \coordinate[right of=B_t_plus] (eq_t_plus);
            \node[clamped, above of=B_t_plus, node distance=15mm] (u_t_plus) {};
            \node[box, below of=eq_t_plus, node distance=20mm] (A_t_plus) {};
            \node[box, below of=A_t_plus, node distance=25mm] (C_t_plus) {};

            \draw[line] (eq_t) -- (B_t_plus);
            \draw[line] (u_t_plus) -- node[anchor=east, pos=0.2]{$u_{t+1}$} (B_t_plus);
            \draw[line] (B_t_plus) -- (eq_t_plus);
            \draw[line] (eq_t_plus) -- node[anchor=east]{$x_{t+1}$} (A_t_plus);
            \draw[line] (A_t_plus) -- node[anchor=east, pos=0.2]{$y_{t+1}$} (C_t_plus);

            \node[clamped, below of=A_t, node distance=12mm] (optim_t) {};
            \node[optim, below of=A_t_plus, node distance=12mm] (optim_t_plus) {$\delta$};

            \msg{up}{right}{D_t_min}{B_t}{0.5}{1}
            \msg{up}{right}{B_t}{eq_t}{0.5}{2}
            \msg{right}{up}{A_t}{C_t}{0.3}{3}
            \msg{right}{up}{A_t}{eq_t}{0.4}{4}
            \msg{up}{right}{eq_t}{B_t_plus}{0.65}{5}

            \draw[dashed] (-0.9, -5.1) rectangle (4.9, 1.9);

            \node (prior) at (0.0,0.8) {$p_t(x_{t-1})$};
            \node (trans_t) at (3.6,1.3) {$p(x_t | x_{t-1}, u_t)$};
            \node (trans_t_plus) at (7.7,0.8) {$p(x_{t+1} | x_t, u_{t+1})$};
            \node (obs_t) at (2.8,-2.0) {$p(y_t | x_t)$};
            \node (obs_t_plus) at (6.4,-2.0) {$p(y_{t+1} | x_{t+1})$};
            \node (goal_t) at (3.0,-4.5) {$\tilde{p}_t(y_t)$};
            \node (goal_t_plus) at (6.6,-4.5) {$\tilde{p}_{t+1}(y_{t+1})$};
        \end{tikzpicture}
    }
    \caption{Message passing schedule for the slide step.}
    \label{fig:slide_schedule}
\end{figure}

\subsection{Results for Interactive Simulation}
\begin{figure}[ht]
    \makebox[\textwidth][c]{
    \begin{tabular}{>{\centering\arraybackslash} m{7cm} >{\centering\arraybackslash} m{7cm}}
        \emph{CBFE Agent} & \emph{EFE Agent}\\
        \includegraphics[width=7cm]{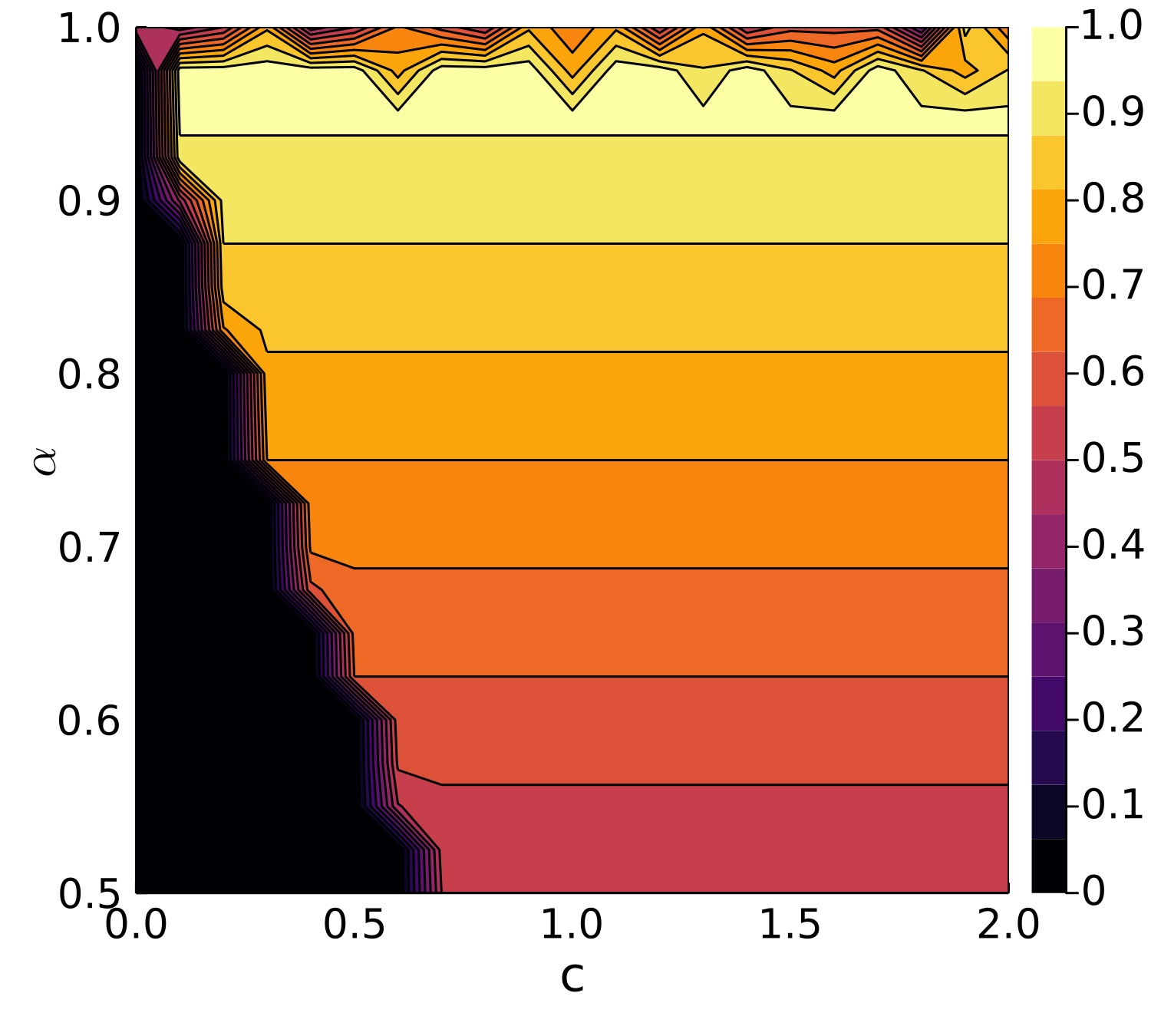} & \includegraphics[width=7cm]{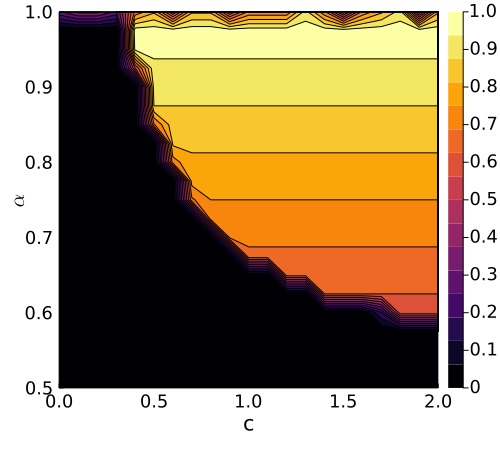}
    \end{tabular}}

    \caption{Average reward landscapes for the Constrained Bethe Free Energy (CBFE) agent and the Expected Free Energy (EFE) agent.}
    \label{fig:reward_results}
\end{figure}

We initialize an environment with the reward in the right arm (position $3$). We then execute the experimental protocol of Alg.~\ref{alg:protocol}, with lookahead $T=2$, for $N=2$ moves, on a dense landscape of varying reward probabilities $\alpha$ and utilities $c$ (scenarios). After the first move, the environment returns an observations to the agent, which informs the agent about second move. After the second move, the expected reward that is associated with the resulting position is reported. We perform $10$ simulations per scenario, and compute the average reward probability. The results of Fig.~\ref{fig:reward_results} compare the average rewards of the CBFE agent and the EFE agent.

From the results of Fig.~\ref{fig:reward_results} it can be seen that the region of zero average reward (dark region in lower left corner) is significantly smaller for the CBFE agent than for the EFE agent. This indicates that the CBFE agent accrues reward in a significantly larger portion of the scenario landscape than the EFE agent. In the lower left corner, the resulting CBFE agent trajectory becomes (4, 4), whereas the EFE agent trajectory becomes (4, 1). Although both agents observe the cue after their first move, they do not visit the indicated reward position in the second move, which leads to zero average reward. Note that neither objective is explicitly designed to optimize for average reward; both define a free energy instead, where multiple simultaneous forces are at play.

\FloatBarrier
In the upper right regions, with high reward probability and utility, both agents consistently execute $(4, 3)$. With this trajectory, the cue is observed after the first move, and the indicated (correct) reward position is visited in the second move, leading to an average reward of $\alpha$. For reward probabilities close to $\alpha = 1$ however, the performance of both agents deteriorates. In this upper region, the informative policies become tied with the greedy policies (see Fig.~\ref{fig:maze_values}), and there is no single dominant trajectory. In some trajectories the agent enters the wrong arm on the first move, from which the agent cannot escape, and the average reward deteriorates.

Greedy behavior is also observed for the CBFE agent when informative priors $c_k$ \eqref{eq:C} are set for all $k$ (including $k=1$), conforming with the configuration of \cite{friston_active_2015}. With this configuration, expected reward for the CBFE agent deteriorates to $0.5$ in the otherwise rewarding region. Interestingly, this change of priors does not affect results for the EFE agent. The resulting change in behavior suggests that the CBFE agent is more susceptible to temporal aspects of the goal prior configuration. While this effect may be considered a nuisance in some cases, it also allows for increased flexibility when assigning explict temporal requirements about goals. For example, assigning an informative versus a flat prior for $k=1$ respectively encodes an urgency in obtaining immediate reward versus a freedom to explore.

\section{Discussion}

In this paper, we focused on epistemic drivers for behavior. We noted that the nature of the epistemic drive differs between an EFE and CBFE agent. Namely, the epistemic drive for the EFE agent stems directly from maximizing a mutual information term between states and observations \eqref{eq:EFE_ep_ex}, while the epistemic drive for the CBFE agent stems from a self-evidencing mechanism (Sec.~\ref{sec:in_ex_val}). In order to better understand the strengths and limitations of the driving forces for the CBFE, it would be interesting to investigate its behavior in more challenging setups, including continuous variables, inference for control \cite{van_de_laar_simulating_2018}, and the effects point-mass constraints on other model variables.

Recent work by \cite{millidge_understanding_2021,da_costa_relationship_2020} shows that epistemic behavior does not occur when the goal prior goes to a point-mass. The work of \cite{millidge_understanding_2021} points to the entropy of the observed variables $\H{q(\bm{y})}$ as a pivotal quantity for epistemic behavior. The CBFE however does not include an entropy over observations, and still exhibits epistemic qualities. The difference in methods lies with the constraint quantity; namely \cite{millidge_understanding_2021,da_costa_active_2020} constrain the goal prior $\tilde{p}(\bm{y}) = \delta(\bm{y} - \hat{\bm{y}})$, while the current paper constrains the variational distribution $q(\bm{y}) = \delta(\bm{y} - \hat{\bm{y}})$ instead. While both constraints remove $\H{q(\bm{y})}$ from the resulting FE objective, optimization of $\hat{\bm{y}}$ in the CBFE still induces an epistemic drive (Sec.~\ref{sec:epistemic_contribution}). Our results thus show that epistemic drives for AIF prove to be more subtle than initially anticipated.

Our presented approach is uniquely scalable, because it employs off-the-shelf message passing algorithms. All message computations are local, which makes our approach naturally amenable to both parallel and on-line processing \cite{bagaev_reactive_2021}. Especially AIF in deep hierarchical models might benefit from the improved computational properties of the CBFE. It will be interesting to investigate how the presented approach generalizes to more demanding (practical) settings.

As a generic variational inference procedure, the CBFE approach applies to arbitrary models. This allows researchers to investigate epistemics in a much wider class of models than previously available. One immediate avenue for further research is the integration of CBFE with predictive coding schemes \cite{bogacz_tutorial_2017,friston_predictive_2009,millidge_predictive_2020}. Predictive coding has so far been driven mainly by minimizing free energy in hierarchical models under the Laplace approximation. Here, the CBFE approach readily applies as well \cite{senoz_variational_2021}, allowing researchers to explore the effects of augmenting existing predictive coding models with epistemic components.

The derivation of alternative functionals that preserve the desirable epistemic behavior of EFE optimization is an active research area \cite{tschantz_scaling_2020,sajid_bayesian_2021}. There have been several interesting proposals such as the Free Energy of the Expected Future \cite{millidge_whence_2020,tschantz_reinforcement_2020,hafner_action_2020} or Generalized Free Energy \cite{parr_generalised_2019}, as well as amortization strategies \cite{millidge_deep_2019,ueltzhoffer_deep_2018} and sophisticated schemes \cite{friston_sophisticated_2021}. Comparing behavior between the CBFE and other free energy objectives might therefore prove an interesting avenue for future research.

In the original description of active inference, a policy precision is optimized during policy planning, and the policy for execution is sampled from a distribution of precision-weighted policies \cite{friston_active_2015}. The present paper does not consider precision optimization, and effectively assumes a large, fixed precision instead. In practice, this procedure consistently selects the policy with minimal free energy; see also \emph{maximum selection} (in terms of value) as described by \cite{schwobel_active_2018}. To accommodate for precision optimization, the CBFE objective might be extended with a temperature parameter, mimicking thermodynamic descriptions of free energy \cite{ortega_thermodynamics_2013}. Optimization of the temperature parameter might then relate to optimization of the policy precision, as often seen in biologically plausible formulations of AIF \cite{fitzgerald_dopamine_2015}.

Another interesting avenue for further research would be the design of a meta-agent that determines the statistics and temporal configuration of the goal priors. In our experiments we design the goal priors \eqref{eq:C} ourselves, such that the agent is free to explore in the first move and seeks reward on the second move. The challenge then becomes to design a synthetic meta-agent that automatically generates an effective lower-level goal sequence from a single higher-level goal definition.

\section{Conclusions}
In this paper we presented mathematical arguments and simulations that show how inclusion of point-mass constraints on the Bethe Free Energy (BFE) leads to epistemic behavior. The thus obtained Constrained Bethe Free Energy (CBFE) has direct connections with formulations of the principle of least action in physics \cite{caticha_entropic_2012}, and can be conveniently optimized by message passing on a graphical representation of the generative model (GM).

Simulations for the T-maze task illustrate how a CBFE agent exhibits an epistemic drive, whereas the BFE agent lacks epistemic qualities. The key intuition behind the working mechanism of the CBFE is that point-mass constraints on observation variables explicitly encode the assumption that the agent will observe in the future. Although the actual value of these observation remains unknown, the agent ``knows'' that it will observe in the future, and it ``knows'' (through the GM) how these (potential) outcomes will influence inferences about states.

We dissected the CBFE objective in terms of its constituent drivers for behavior. In the CBFE framework, in addition to being functionals of the state beliefs, the confidence and complexity are viewed as functions of the potential outcomes and policy respectively. Simultaneous optimization of variational distributions and potential outcomes then leads the agent to prefer epistemic policies. Interactive simulations for the T-maze showed that, compared to an EFE agent, the CBFE agent incurs expected reward in a significantly larger portion of the scenario landscape.

We performed our simulations by message passing on a Forney-style factor graph representation of the generative model. The modularity of the graphical representation allows for flexible model search, and message passing allows for distributed computations that scale well to bigger models. Constraining the BFE and optimizing the CBFE objective by message passing thus suggests a simple and general mechanism for epistemic-aware AIF in free-form generative models.

\section*{Conflict of Interest Statement}
The authors declare that the research was conducted in the absence of any commercial or financial relationships that could be construed as a potential conflict of interest.

\section*{Author Contributions}
The original idea was conceived by TvdL. All authors contributed to further conceptual development of the methods that are presented in this manuscript. Simulations were performed by TvdL and MK. All authors contributed to writing the manuscript.

\section*{Funding}
This research was made possible by funding from GN Hearing A/S. This work is part of the research programme Efficient Deep Learning with project number P16-25 project 5, which is (partly) financed by the Netherlands Organisation for Scientific Research (NWO).

\section*{Acknowledgments}
The authors gratefully acknowledge stimulating discussions with Dimitrije Markovi\'{c} of the Neuroimaging group at TU Dresden. The authors also thank the two reviewers for their valuable comments.

\section*{Abbreviations}
The following abbreviations are used in the manuscript:

\begin{tabular}{l l}
FEP  & Free Energy Principle\\
AIF  & Active Inference\\
VFE  & Variational Free Energy\\
EFE  & Expected Free Energy\\
BFE  & Bethe Free Energy\\
CBFE & Constrained Bethe Free Energy\\
GM   & Generative Model\\
EM   & Expectation Maximization\\
FFG  & Forney-style Factor Graph\\
VMP  & Variational Message Passing\\
SP   & Sum-Product\\
MAP  & Maximum A-Posteriori
\end{tabular}

\bibliographystyle{ieeetr}
\bibliography{bibliography.bib}

\appendix
\section*{Appendix}
\subsection*{A. Evaluation of the Expected Free Energy}

The standard procedure for evaluating the Expected Free Energy (EFE) collects instantaneous EFE contributions over time by a forward filtering approach \cite{friston_active_2015}. Following \cite{friston_active_2015,da_costa_active_2020}, the EFE constructs an instantaneous model for each future time-point $\tau \geq t$, as
\begin{align}
    f(y_{\tau}, x_{\tau} | \bm{u}_{t:\tau}) &= p(x_{\tau} | y_{\tau}, \bm{u}_{t:\tau})\, \tilde{p}(y_{\tau})\,, \label{eq:gen-model-EFE}
\end{align}
with $\tilde{p}(y_{\tau})$ the goal prior, and $p(x_{\tau} | y_{\tau}, \bm{u}_{t:\tau})$ a state posterior that needs to be further defined.

Using Bayes rule, we can express the state posterior in terms of the observation model and a posterior predictive for the state, as
\begin{subequations}\label{eq:p-t-y-t}
\begin{align}
    p(x_{\tau} | y_{\tau}, \bm{u}_{t:\tau}) &= \frac{p(x_{\tau} | \bm{u}_{t:\tau}) p(y_{\tau} | x_{\tau})}{p(y_{\tau} | \bm{u}_{t:\tau})} \label{eq:EFE_GM_Bayes}\\
    &= \frac{p(x_{\tau} | \bm{u}_{t:\tau}) p(y_{\tau} | x_{\tau})}{\sum_{x_{\tau}} p(x_{\tau} | \bm{u}_{t:\tau}) p(y_{\tau} | x_{\tau})}\,. \label{eq:EFE_GM_known_quantities}
\end{align}
\end{subequations}

The posterior predictive $p(x_{\tau} | \bm{u}_{t:\tau})$ is explicitly conditioned on the policy $\bm{u}_{t:\tau}$, from current time $t$ up to and including future time $\tau$, and thus represents the forward prediction (filtering solution) for the current state belief given preceding controls (whilst excluding preceding goals). Using the generative model engine definition, the posterior predictive for the state then becomes\footnote{The definition of \cite[p.~192]{friston_active_2015} implicitly defines this forward prediction as a marginalization over states, which is made explicit in the definition of \eqref{eq:EFE_filtering}. In general, this marginalization need not be tractable, in which case it can also be approximated by on-line optimization of an appropriate BFE objective on the generative model engine.}
\begin{subequations}
\label{eq:EFE_filtering}
\begin{align}
    p(x_{\tau} | \bm{u}_{t:\tau}) &= \sum_{\bm{y}_{t:\tau}} \sum_{\bm{x}_{t-1:\tau-1}} p(x_{t-1}) \prod_{k=t}^{\tau} p(y_k, x_k | x_{k-1}, u_k)\\
    &= \sum_{\bm{x}_{t-1:\tau-1}} p(x_{t-1}) \prod_{k=t}^{\tau} p(x_k | x_{k-1}, u_k)\,.
\end{align}
\end{subequations}
The second step of \eqref{eq:EFE_filtering} simplifies the expression by marginalizing over $\bm{y}_{t:\tau}$. The posterior predictive can then conveniently be computed by message passing on the generative model \cite{blahut_least_2002}, using a single forward pass.

We are now prepared to construct the instantaneous EFE (the EFE at time $\tau$), which is defined as \cite{friston_active_2015}
\begin{align}
    \G_{\tau}(\hat{\bm{u}}_{t:{\tau}}) = \E{p(y_{\tau} | x_{\tau})\,p(x_{\tau} | \hat{\bm{u}}_{t:\tau})}{\log \frac{p(x_{\tau} | \hat{\bm{u}}_{t:\tau})}{f(y_{\tau}, x_{\tau} | \hat{\bm{u}}_{t:\tau})}}\,. \label{eq:EFE_instantaneous}
\end{align}

Upon substitution of \eqref{eq:EFE_GM_Bayes} in \eqref{eq:EFE_instantaneous}, the instantaneous EFE factorizes into ambiguity and risk, as
\begin{align}\label{eq:EFE_risk_ambiguity}
    \G_{\tau}(\hat{\bm{u}}_{t:{\tau}}) &= \E{p(y_{\tau} | x_{\tau})\,p(x_{\tau} | \hat{\bm{u}}_{t:\tau})}{\log \frac{p(y_{\tau} | \hat{\bm{u}}_{t:\tau})}{p(y_{\tau} | x_{\tau})\,\tilde{p}(y_{\tau})}} \notag\\
    &= -\E{p(x_{\tau} | \hat{\bm{u}}_{t:\tau})}{\E{p(y_{\tau} | x_{\tau})}{\log p(y_{\tau} | x_{\tau})}} + \E{p(y_{\tau} | x_{\tau})\,p(x_{\tau} | \hat{\bm{u}}_{t:\tau})}{\log \frac{p(y_{\tau} | \hat{\bm{u}}_{t:\tau})}{\tilde{p}(y_{\tau})}}\notag\\
    &= \underbrace{\E{p(x_{\tau} | \hat{\bm{u}}_{t:\tau})}{\H{p(y_{\tau} | x_{\tau}) }}}_{\text{ambiguity}} + \underbrace{\KL{p(y_{\tau} | \hat{\bm{u}}_{t:\tau})}{\tilde{p}(y_{\tau})}}_{\text{observation risk}}\,.
\end{align}
This decomposition is often used to compute the instantaneous EFE in practice.

The complete EFE of the full policy $\hat{\bm{u}}$ then follows by summation of all instantaneous contributions
\begin{align}
    \G(\hat{\bm{u}}) = \sum_{\tau=t}^{t+T-1} \G_{\tau}(\hat{\bm{u}}_{t:\tau})\,. \label{eq:EFE}
\end{align}

To summarize, the procedure for computation of the EFE in practice \cite{friston_active_2015,da_costa_active_2020} usually consists of three steps. First, for a given policy $\hat{\bm{u}}$, the posterior predictive distributions \eqref{eq:EFE_filtering} are computed for all $t \leq \tau < t+T$. Then, the instantaneous EFE's are (individually) computed. Finally, the instantaneous EFE's are summed to produce the full-policy EFE \eqref{eq:EFE}.

\end{document}